\documentclass{styles/SVProc}
\usepackage{makeidx}  
\makeindex
\usepackage{url}
\usepackage[noadjust]{cite}

\usepackage{booktabs}
\usepackage{subfigure}
\usepackage{makecell}
\usepackage{graphicx}
\usepackage{amsmath, amsfonts, amssymb}

\begin{document}
 \frontmatter          
\pagestyle{headings}  
\title{Velocity and stroke rate reconstruction of canoe sprint team boats based on panned and zoomed video recordings}
\titlerunning{Reconstruction of canoe velocity and stroke rate}  

\newif\ifreview
\reviewfalse

\ifreview
  \author{Anonymous Authors}
  \institute{Paper ID: 50}
\else

\author{Julian Ziegler$^{1, *}$ \and Daniel Matthes$^{1, *}$ \and Finn Gerdts$^{1}$ \and Patrick Frenzel$^{1}$ \and Torsten Warnke$^{2}$ \and Matthias Englert$^{2}$ \and Tina Kövari$^{3}$ \and Mirco Fuchs$^{1}$}
\authorrunning{Julian Ziegler, Daniel Matthes et al.} 
%
%
\index{Ziegler, J.}
\index{Matthes, D.}
\index{Gerdts, F.}
\index{Frenzel, P.}
\index{Warnke, T.}
\index{Englert, M.}
\index{Kövari, T.}
\index{Fuchs, M.}

\institute{Laboratory for Biosignal Processing, Leipzig University of Applied Sciences, Leipzig, Germany\\
$^*$ Equal Contributions\\
\and
Research Group Canoeing, Institute for Applied Training Science (IAT), Leipzig, Germany\\
\and
German Canoe Federation, Duisburg, Germany}

\fi
\maketitle

\begin{abstract}
Pacing strategies, defined by velocity and stroke rate profiles, are essential for peak performance in canoe sprint. While GPS is the gold standard for analysis, its limited availability necessitates automated video-based solutions. This paper presents an extended framework for reconstructing performance metrics from panned and zoomed video recordings across all sprint disciplines (K1–K4, C1–C2) and distances (200m–500m). Our method utilizes YOLOv8 for buoy and athlete detection, leveraging the known buoy grid to estimate homographies. We generalized the estimation of the boat position by means of learning a boat-specific athlete offset using a U-net based boat tip calibration. Further, we implement a robust tracking scheme using optical flow to adapt to multi-athlete boat types. Finally, we introduce methods to extract stroke rate information from either pose estimations or the athlete bounding boxes themselves.
Evaluation against GPS data from elite competitions yields a velocity \textit{MAPE} of 0.011 [0.008 0.014] (Spearman $\rho=0.974$) and a stroke rate \textit{MAPE} of 0.009 [0.006 0.013] (Spearman $\rho = 0.975$). The methods provide coaches with highly accurate, automated feedback with minimal manual initialization work required, and without requiring sensors.
\end{abstract}

\section{Introduction}

Pacing strategies are fundamental to achieve top performance in canoe sprint. An elite performance is characterized by the complex interplay between velocity profiles and stroke rate profiles, which indicate how velocity and stroke rate, respectively, evolve over the course of the race. To gain a competitive advantage, coaches and sports scientists require precise, high-frequency data to analyse tactical execution of all competitors throughout a race. While Global Positioning System (GPS) technology is currently considered the gold standard for acquiring such data~\cite{fernandes_validation_2024}, its practical utility is often constrained. In high-stakes competition, GPS data is rarely available for all participants, and the deployment of on-boat sensors can be logistically cumbersome or prohibited by regulations. 

As an alternative, video-based analysis offers a non-invasive means of data collection. However, manual video analysis using panned and zoomed recordings from the shore is frequently deemed unfeasible. The skewed perspective inherent in side-view filming renders the manual reconstruction of absolute boat positions extremely difficult and prohibitively labour-intensive. Consequently, there is a significant need for automated systems capable of transforming standard broadcast or coach-recorded footage into accurate kinematic data.

\begin{figure}
    \centering
    \includegraphics[width=1.0\linewidth]{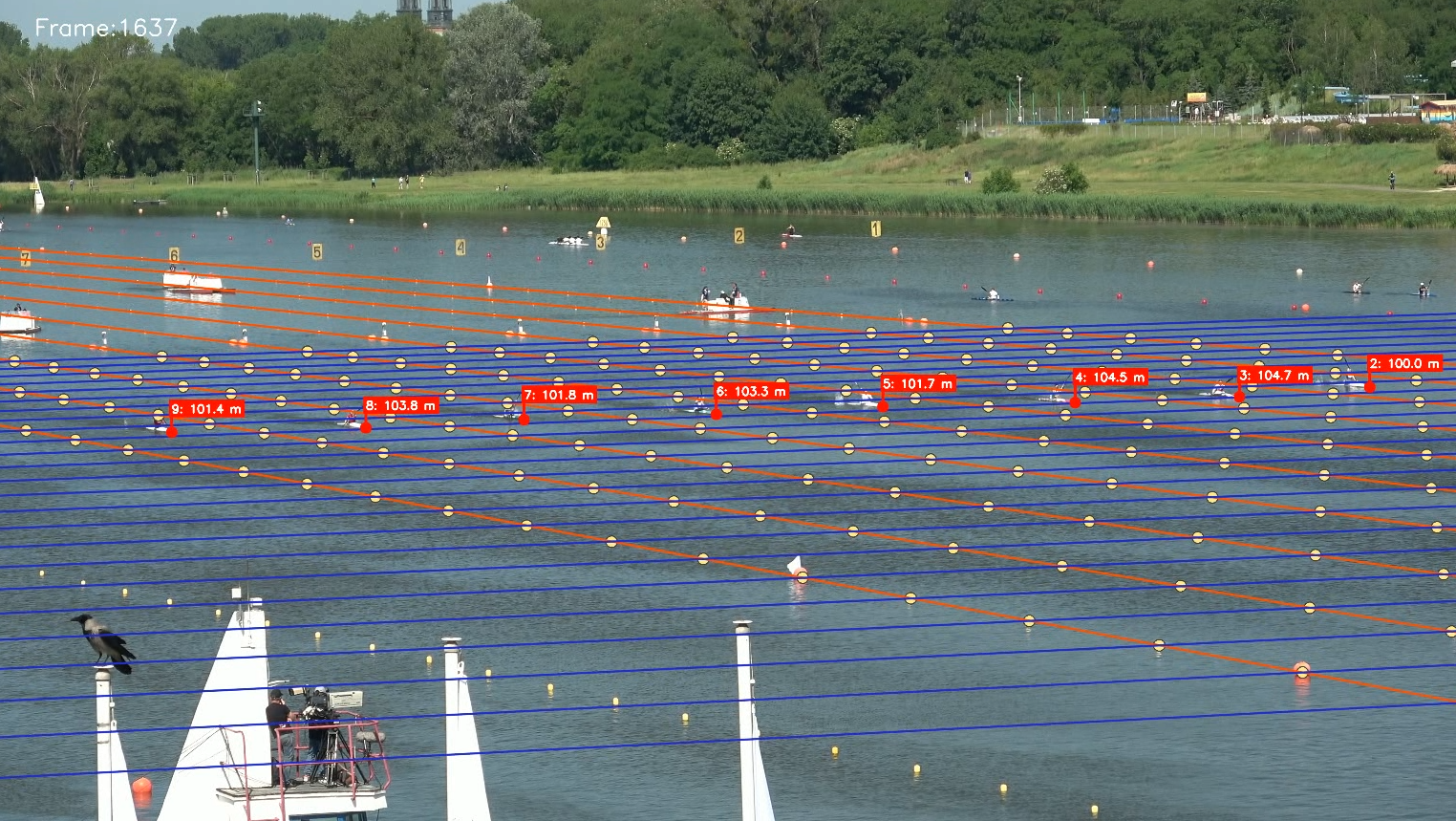}
    \caption{Visualization of the reconstructed scene geometry for a K1 canoe race over the distance of 500~m according to~\cite{matthes_reconstructing_2025}. Orange lines correspond to lane boundaries, blue ones to boundaries between equidistant segment, and yellowish dots where orange and blue lines cross are expected buoy location alongside the track. See Sec. \ref{sec:prevwork} for an outline of the scene reconstruction method. Numbers in red coloured annotations at boat tips correspond to the distance travelled of each boat, which can further be utilized to reconstruct velocity profiles. Best viewed on screen.}
    \label{fig:K1_01}
\end{figure}

In previous work, Matthes et al. introduced a method to automatically reconstruct velocity profiles for single-person boats (K1/C1) using a fixed camera position \cite{matthes_reconstructing_2025}. This approach utilized a YOLOv8~\cite{varghese_yolov8_2024} object detection model to identify buoys and athletes, leveraging the known geometry of the buoy grid to estimate homographies and establishing buoy correspondences between frames via optical flow (cmp. Fig.~\ref{fig:K1_01}). While this method proved highly accurate for 500m single-person races, it was limited in scope to individual disciplines and did not provide means for stroke rate reconstruction.

This paper significantly extends that framework in three key areas, and expands the methodology to team boats (K2, C2, K4).
We first introduce an automated calibration to determine the athlete offset used to localize the boat on a per video, per lane basis. To achieve this, a U-Net model is trained to detect boat tips.
The multi-athlete classes introduce significant complexity due to potential detection failures among multiple athletes.
To resolve this, we implement a robust tracking logic that identifies frames with complete detections and utilizes optical flow to maintain athlete order, as well as position estimates in adjacent frames.
Third, we incorporate stroke rate extraction into the pipeline. Two methods of capturing a one-dimensional motion signal are introduced, either using a ViTPose~\cite{xu_vitpose_2022} model or the YOLO bounding boxes. From the motion signal, we derive stroke rate.

A comprehensive evaluation of the system across a diverse range of distances (200m, 500m) and boat types is provided. The resulting framework delivers a robust, automated, camera-based system for tactical race analysis, providing immediate, actionable feedback across all canoe sprint disciplines without the need for on-boat hardware.

\section{Related Works}
Exploiting static scene geometry to establish correspondences between image observations and real-world coordinates is a well-established strategy in computer vision. When parts of the environment provide reliable geometric structure (e.g., planes, lines, or repeated landmarks), these constraints can be used to infer camera parameters and spatial mappings with reduced dependence on manual annotation or specialised sensors. Such approaches have been applied successfully to problems including camera calibration, object localisation, and scene understanding~\cite{nishida_road_2011, taniai_continuous_2018}.

Sport analytics is a particularly strong application domain for geometry-based correspondence, because the environment often contains fixed markings and largely standardised camera viewpoints. In field sports, line markings and known layout dimensions enable estimation of homographies or full camera models, which in turn support metric player tracking and the extraction of spatiotemporal performance measures from broadcast or consumer-grade footage~\cite{yagi_estimation_2018, gutierrez-perez_no_2024, claasen_video-based_2024, ziegler_auxflow_2026}. By anchoring visual measurements to a calibrated world coordinate frame, these methods provide scalable alternatives to labour-intensive per-frame annotation and facilitate robust analysis under realistic capture conditions.

The biomechanical analysis of flatwater kayaking and canoeing has a long-standing history of relying on visual data for performance evaluation. Early foundational studies by Plagenhoef~\cite{plagenhoef_biomechanical_1979} and Mann and Kearney~\cite{mann_biomechanical_1980} utilized slow-motion cinematography to extract kinematic and kinetic parameters, establishing a basis for differentiating between elite and sub-elite paddling techniques. As highlighted by Michael et al.~\cite{michael_determinants_2009}, video- and film-supported analyses have remained a central determinant in evaluating kayak performance. Over time, these visual methods evolved into highly structured observational models; for example, McDonnell et al.~\cite{mcdonnell_observational_2012} utilized multi-camera video recordings to explicitly define and assess visible phase positions in sprint kayaking technique.

More recently, traditional video analysis has been combined with modern sensor technologies to provide a more comprehensive biomechanical picture. Side-view video recordings frequently serve as a critical ground-truth reference to validate wearable GPS and accelerometer data for metrics like speed and stroke rate~\cite{fernandes_validation_2024}. Similarly, Romagnoli et al.~\cite{romagnoli_paddle_2025} successfully combined 2D video analysis with instrumented paddles and GPS tracking to evaluate propulsive force and power balance.

In competitive canoeing and related water sports, recent work has begun to adapt modern deep learning and computer vision principles to aquatic environments, where reflections, spray, and limited stable reference structure make calibration and measurement more challenging. Prior studies have leveraged waterline segmentation as a persistent visual reference for technique assessment~\cite{von_braun_utilizing_2020}, and demonstrated the feasibility of video-based metric extraction for performance evaluation~\cite{bonaiuto_system_2022}. Beyond 2D analysis, monocular reconstruction of the full 3D paddle motion has also been explored~\cite{najlaoui_ai-driven_2024}, indicating that meaningful kinematic descriptors can be obtained even without multi-camera setups. These developments align naturally with advances in temporal 3D pose estimation~\cite{pavllo_3d_2019} and learning-based camera calibration pipelines~\cite{sha_end--end_2020}, which together reduce reliance on controlled setups and enable increasingly automated analysis.

Complementary lines of research address higher-level race and crew dynamics, such as the quantification of stroke synchronisation~\cite{tay_video-based_2018}, the analysis of pacing profiles~\cite{estreich_analysis_2025}, and broader algorithm-based approaches for competitive canoeing performance analysis~\cite{amat_algorithm-based_2025}.

\section{Methodology}

In this section, we present our proposed methodology. The framework is designed specifically to address the limitations identified in prior work, particularly regarding multi-athlete boat types. A synopsis of this work, concerning single-athlete velocity estimation is given in the first section.
The primary challenge in these environments is the increased complexity of spatial association and the necessity for high-precision geometric calibration that accounts for the unique configuration of team boats.

To overcome these hurdles, we first introduce the general mathematical framework and adjust notations to also handle team boats. Second, we present an extended dataset, which includes the multi-athlete boat types K2, C2, K4, as well as more C1 and K1 videos than the original dataset.
Moreover, a boat tip dataset to train the U-Net was also curated.

We then detail our automatic boat tip calibration, enabling superior boat localization compared to using a static, empirical offset. 
To this end, a U-Net model was implemented to detect the boat tip given an image patch, which we feature in section \ref{sec:meth-unet}.
Further, we provide a description of our multi-athlete tracking using optical flow, enabling athlete order disambiguation and positional tracking in the event of failed detections or occlusions.

\subsection{Single-Athlete Homography-based Velocity Estimation}
\label{sec:prevwork}

\begin{figure}
    \centering
    \includegraphics[width=\linewidth]{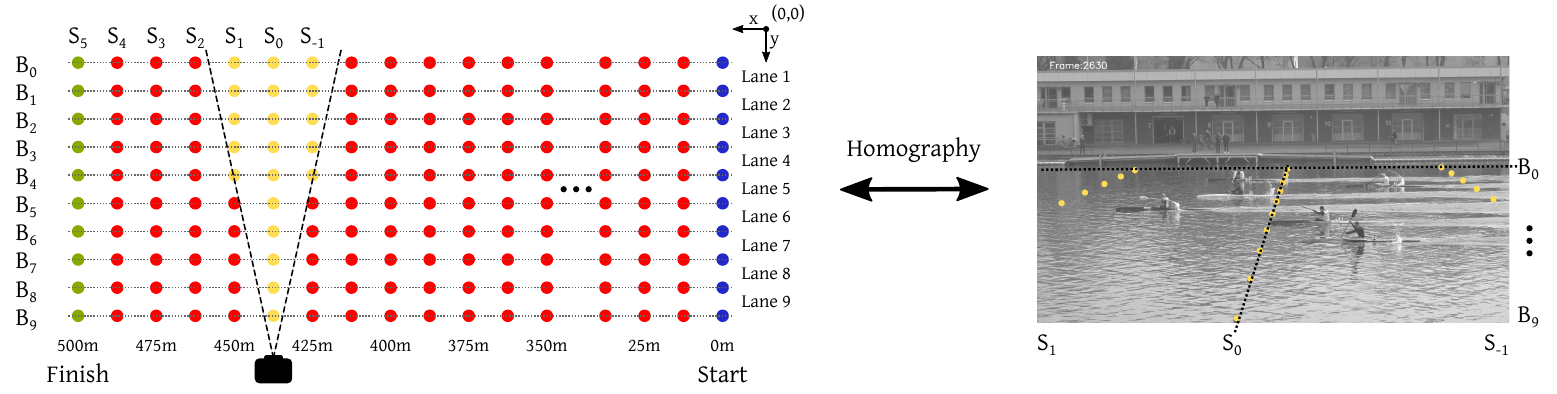}
    \caption{Shown is the scene geometry of a regatta course (left) and an approximate orthogonal camera view on the scene (right). For clarity, yellow buoys
highlighted in the scene are those visible in the right image. Corresponding buoys in both views can be identified based on their lane boundaries $B_i$ and track
segments $S_j$. Correspondences are used to estimate a homography and propagated to adjacent frames. Adapted from~\cite{matthes_reconstructing_2025}.}
    \label{fig:orig-method}
\end{figure}

This work builds on earlier methods for reconstructing boat velocity profiles from race video~\cite{matthes_reconstructing_2025}. The original approach exploits the standardised geometry of a regatta course to turn image measurements into metric positions on the planar water surface (Fig.~\ref{fig:orig-method}). Buoys and athletes are detected in each frame using a fine-tuned YOLOv8 model, and the fixed buoy layout provides reliable reference points to estimate an image-to-course mapping. Because correspondences are easiest to initialise in a near-orthogonal view, the mapping is first set up in a frame where lane boundaries appear approximately horizontal, fitted robustly (RANSAC), and then carried through the full sequence by tracking buoy locations with Lucas-Kanade optical flow. Propagating the mapping forward to the finish and backward to the start yields a consistent track coordinate frame for every frame, even when buoy detections are intermittent due to occlusions or missed detections.

Boat position is approximated indirectly from the detected athlete location by using a point near the bottom of the athlete bounding box as a proxy for where the boat lies in the image. The resulting trajectory is smoothed to reduce jitter from detection noise and athlete motion, velocity is obtained by differentiating the smoothed position over time, and the final profiles are reported as averages over fixed 12.5\,m segments in line with common coaching practice.

Despite its effectiveness for single-athlete boats, the method has important limitations that motivate the present work. Most notably, it does not extend cleanly to crew boats, because multiple athletes are visible at once and often overlap; the detection that serves as a reliable proxy in a single-athlete setting becomes ambiguous when several athletes could plausibly define the boat position. In addition, the boat tip is not observed directly and is instead approximated by adding a constant empirical offset (2.4\,m, tuned to German boat measurements) to the mapped athlete position, which can introduce a systematic bias when boat class, athlete placement, or recording conditions differ. This issue is amplified for team boats, where the correct offset is no longer global: it depends on which seat (and thus which athlete) is used as the proxy, making a single correction insufficient. Finally, the pipeline focuses on velocity profiles and does not provide stroke profile information.

\subsection{Expansion to Multi-Athlete Races}

First, we expand the method to handle multi-person boats and update the notation accordingly. Consistent with previous work, we estimate a homography $\mathbf{H}_i$ for each frame $f_i$, which enables the mapping of image points on the water surface plane to two-dimensional real-world coordinates.

We utilize a trained YOLOv8 model to detect athletes in every frame. The set of detections per frame is denoted as $\mathcal{B}_i'$, where the prime symbol ($'$) indicates vectors or sets residing in the image space. The 2d image position of an athlete $\mathbf{a}'$ is defined as the centre of the bottom edge of their respective bounding box $\mathbf{b}' \in \mathcal{B}_i'$, resulting in the set of athlete image positions $\mathcal{A}'_i$.

Given the following transformation:
\begin{equation}
    w \begin{pmatrix} \mathbf{a} \\ 1 \end{pmatrix} = \mathbf{H}_i \cdot \begin{pmatrix} \mathbf{a}' \\ 1 \end{pmatrix}
\end{equation}
we obtain the 2d positions of each athlete in world coordinates (i.e., a position on the water surface) and therefore a set of athlete world positions $\mathcal{A}_i$ per frame, where $w$ is the homogeneous scaling factor. 

By incorporating knowledge of lane boundaries, we can partition these detections into lane-specific subsets $\mathcal{A}_{i,j}$, where $j$ represents the lane number. For a given lane $j$ and frame $f_i$, the set $\mathcal{A}_{i,j}$ contains $N$ elements, depending on the boat type (e.g., K1, K2, or K4). Each individual athlete position is denoted as $\mathbf{a}_{i,j,k}$, with $k \in \{1, \dots, N\}$. These positions are ordered according to the boat's direction of travel, i.e. the first athlete is the one closest to the tip of the boat.

Because the position of a boat in canoe sprint races is defined by its tip, we apply a set of correction offsets $\mathbf{o}_{j,k}$ from each athlete $k$ to the boat tip. The estimation of these offsets is detailed in Section~\ref{sec:meth-cal}. Finally, the boat position $\mathbf{x}_{i,j}$ in frame $f_i$ and lane $j$ in real-world coordinates is computed as:
\begin{equation}
    \mathbf{x}_{i,j} = \frac{1}{N} \sum_{k=1}^{N} (\mathbf{a}_{i,j,k} + \mathbf{o}_{j,k})
\end{equation}
Evaluating these coordinates over time $t$ yields a lane-specific distance-time signal $\mathbf{x}_j(t)$, the further application of which is discussed in the subsequent sections.

\begin{figure}
    \centering
    \includegraphics[width=1.0\linewidth]{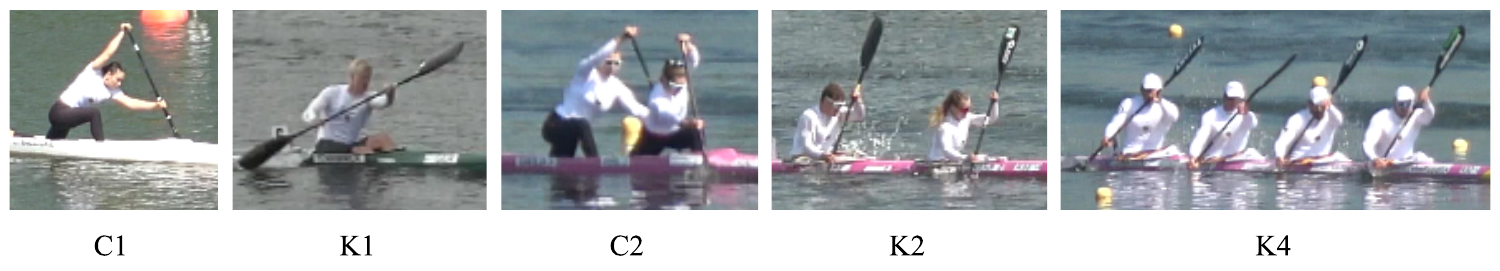}
    \caption{Visualization of all boat classes in our main dataset. C refers to canoe, where single-bladed paddles are used. K refers to kayak, where double-bladed paddles are mandated. The number indicated the number of athletes per boat.}
    \label{fig:classes}
\end{figure}

\subsection{Datasets}
This section presents the datasets used throughout this study. We distinguish between a main dataset and a curated dataset that is only used to train the boat tip detection model. Both are datasets are outlined below.

The main dataset comprises 5 boat types, single-seated C1 and K1, double seated C2 and K2, and quadruple seated K4, where C refers to a canoe (single bladed paddle) and K to a kayak (double bladed paddle). See figure \ref{fig:classes} for a sample image of each boat type.
The video recordings were obtained from the German National Qualifiers 2023 in Duisburg, the 2023 ICF Canoe Sprint World Championships in Duisburg, and the 2024 ICF Canoe Sprint World Cup in Poznan, Poland. We captured at 4K resolution and 25 fps from an elevated position 40-90~m before the finish line.
GPS data was acquired by GPS modules mounted at the stern, and data post-processed to align trajectories to a common starting reference. Gyroscopic stroke rate data is processed in the same way. Overall, the main dataset contains 54 GPS and Gyro tracks across 40 race video recordings, as some videos include multiple boats for which GPS and Gyro data is available. We include both 200~m and 500~m races.
Table~\ref{tabl:Dataset} shows the exact split across all boat types and race distances.
Following Fernandes et al.~\cite{fernandes_validation_2024}, we consider the sensor data to be a good ground truth, although disagreement in the first 50~m can be the result of observed errors in the GPS measurement.

\begin{table}[h]
\centering
\label{tabl:Dataset}
\caption{Contents of our main dataset.}
\renewcommand{\arraystretch}{1.2}
\begin{tabular}{l|cc|cc|c|c}
\hline
\textbf{Distance} & \textbf{K1} & \textbf{C1} & \textbf{K2} & \textbf{C2} & \textbf{K4} & \textbf{Total} \\ \hline
200 m             & 8           & 3           & --          & --          & --          & 11             \\
500 m             & 21          & 6           & 7           & 5           & 4           & 43             \\
\textbf{Total}    & 29          & 9           & 7           & 5           & 4           & 54          \\\hline
\end{tabular}
\end{table}

Furthermore, we build a boat tip dataset of image patches with a $150 \times 150$ pixel resolution. 
The patches were selected according to the method described in Sec. \ref{sec:meth-cal}, resulting in patches where the boat tip is near the centre coordinate of the patch.
In each patch, the boat tip is manually annotated. The resulting dataset comprises 550 train samples, 124 validation datapoints and 98 testcases.

\subsection{Automatic Boat Tip Offset Calibration}
\label{sec:meth-cal}

In this section, we detail the derivation of the correction offsets $\mathbf{o}_{j,k}$, which are applied to athlete positions to localize the boat. While the previous work focused on single-seater vessels using a single empirical offset, we now introduce a method to automatically calibrate all $N$ offsets for a boat in lane $j$.

\begin{figure}
    \centering
    \includegraphics[width=1.0\linewidth]{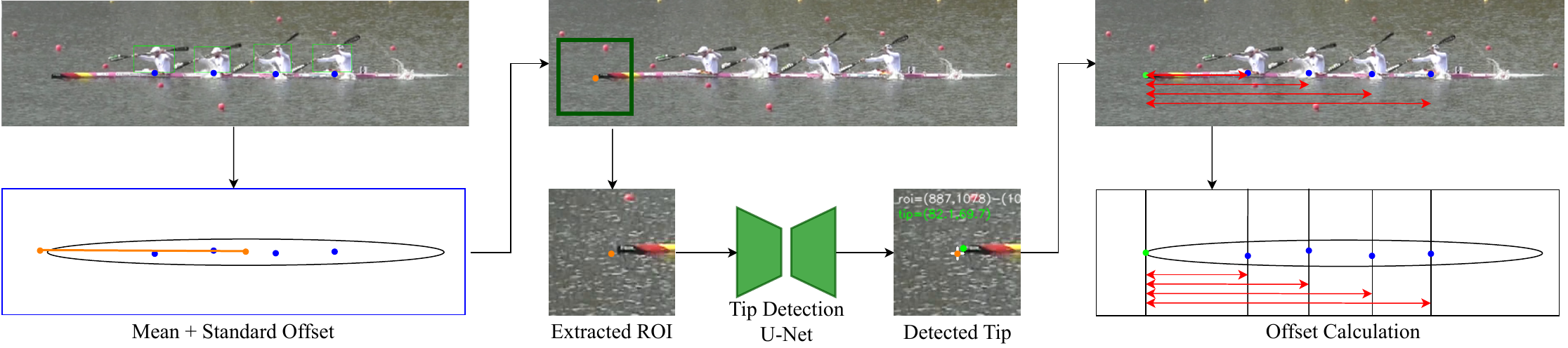}
    \caption{Graphical Illustration of Automatic Boat Tip Offset Calibration. (1) For every detected athlete bounding box (green), we assume the centre of the bottom edge to be the image position of that athlete (blue point). (2) Athlete positions are transformed to the 2d world space, and a standard offset is applied to the mean position of the athletes (orange). (3) This position is reprojected into the image, and defined as centre of ROI. (4) ROI is input into a trained U-Net, which predicts the exact tip location (bright green). (5) Exact tip location and athlete positions in the image are again transformed to world coordinates. (6) Individual athlete offsets are calculated (red arrows). Best viewed on screen.}
    \label{fig:offset}
\end{figure}

To achieve this, a sequence of five frames is selected from a near-perpendicular view where all athletes are detected. For each frame $i$, we follow the pipeline illustrated in Fig. \ref{fig:offset}. First, an initial estimate of the boat position in world coordinates, $\tilde{\mathbf{p}}_{i,j}$, is calculated by applying a boat-type specific empirical offset $ \mathbf{o}(N)$ to the mean of the athlete positions $\mathcal{A}_{i,j}$:

\begin{equation}
    \tilde{\mathbf{p}}_{i,j} = \mathbf{o}(N) +  \frac{1}{N}\sum_{k=1}^N\mathbf{a}_{i,j,k}.
\end{equation}

The corresponding image position $\tilde{\mathbf{p}}'_{i,j}$ serves as the centre pixel for a $150 \times 150$ pixel Region of Interest (ROI), i.e. a patch that is expected to contain the true boat tip. This ROI is processed by a U-Net model specifically trained for boat tip detection (see Sec. \ref{sec:meth-unet}). The model outputs the precise image location of the boat tip, $\mathbf{p}'_{i,j}$, which is then transformed back into world coordinates $\mathbf{p}_{i,j}$ using the homography matrix $\mathbf{H}_{i}$. 

Consequently, the individual athlete offsets are derived as:

\begin{equation}
    \mathbf{o}_{j,k} = \mathbf{p}_{i,j} - \mathbf{a}_{i,j,k}.
\end{equation}

To ensure temporal robustness and mitigate detection noise, these offsets are averaged over the five-frame sequence to produce the final correction offsets.

\subsection{U-Net model for Boat Tip Detection}
\label{sec:meth-unet}

\begin{figure}
    \centering
    \includegraphics[width=1.0\linewidth]{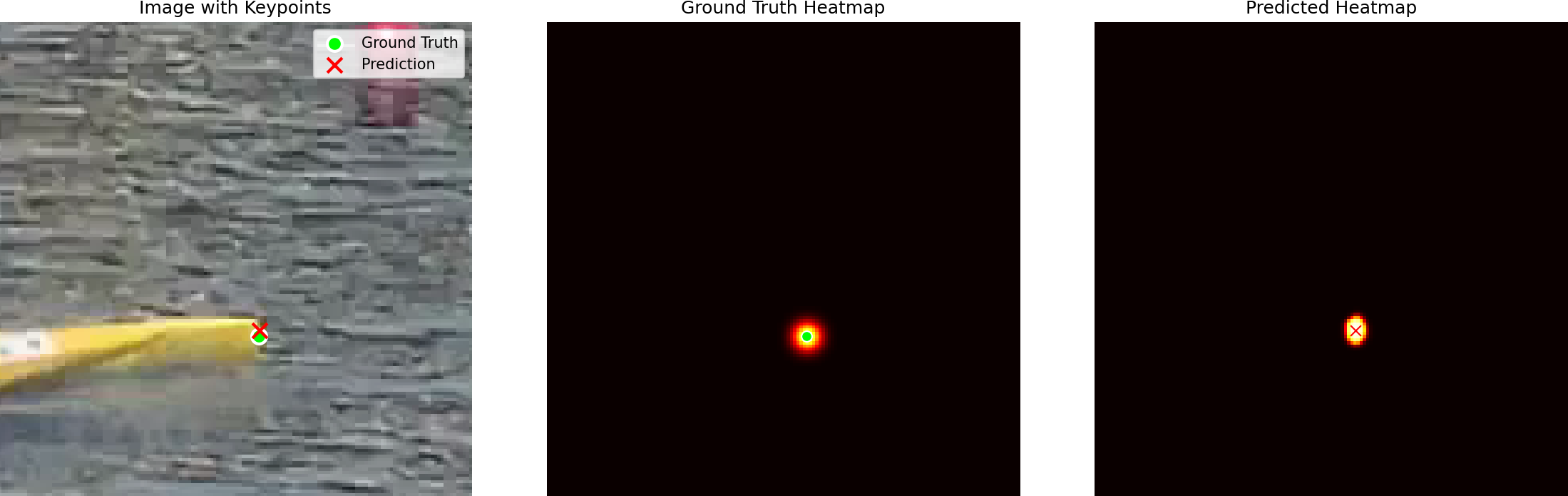}
    \caption{Principle of boat tip localization using a U-Net. The image annotation (green dot) of the boat tip (left) is encoded via a heatmap with gaussian kernel (middle). The U-Net is trained to predict this heatmap (right) from which the predicted position (red cross) is derived using argmax over the predicted heatmap.}
    \label{fig:unet-heatmap}
\end{figure}

We aim to learn a U-Net model to predict the boat tip present in a $150 \times 150$ pixel patch, i.e. regressing the position of the tip.
For this, we formulate this problem as a classification of all pixels to either be or not be the boat tip. 
The target of the U-Net is a heatmap built from the ground truth boat tip position and a Gaussian kernel with a specified $\sigma$. 
This approach is common in other key point detection tasks such as pose estimation \cite{xu_vitpose_2022} or sports field registration \cite{gutierrez-perez_no_2024}.The principle is visualized in Fig. \ref{fig:unet-heatmap}.

\subsubsection{Model Architecture}

The neural network utilizes a ResNet-18 backbone, pretrained on ImageNet, as its encoder. This stage is responsible for generating deep, low-resolution feature maps that identify the existence and approximate spatial placement of keypoints. To maintain the integrity of the spatial structure during feature extraction, Groupnorm is implemented. 

The custom decoder restores spatial resolution through successive upsampling stages. A critical component of this architecture is the use of skip connections, which transfer low-level feature information directly from the encoder to the decoder. This mechanism facilitates a significantly more precise localization of the features identified in the earlier layers. The decoder consists of convolutional layers with ReLU activation and concludes with an output head that generates a heatmap of dimensions $(1, H, W)$.

\subsubsection{Training and Optimization}

Training is conducted using a dual-loss strategy to maximize localization accuracy. Binary Cross Entropy (BCE) loss is applied to the heatmaps to identify probable pixel locations, while Mean Squared Error (MSE) loss is utilized on the final coordinates to further refine the predictions.

The optimization process employs a Onecycle scheduler, with the decoder's learning rate set ten times higher than that of the encoder. Furthermore, a sigma curriculum is applied to the dataset heatmaps, where the value of $\sigma$ is reduced from $3$ to $1.5$ throughout the training duration. This curriculum forces the model to achieve increasingly accurate localization as the training progresses.

\subsection{Robust Multi-Athlete Boat Localization}
\label{sec:meth-multi}

Until now, we have only considered frames in which the number of detected athletes per lane, i.e. $|\mathcal{A}_{i,j}|$, equals the number of athletes $N$ in the respective boat type.
This, however, is of course not always the case.
Furthermore, if any of the $N$ athletes is failed to be detected, not only is that athlete's position unknown, it makes it impossible to infer the correct positional order of all other athletes. This, in turn, results in the correction offsets to not be able to be applied.

Given the set of all frames $\mathcal{F}$, there exist lane specific subsets $\mathcal{F}^c_j$ where all athletes of that lane are correctly detected in the frames within this subset.

Without further handling, the distance-time-signal $\textbf{x}_j\left(t\right)$ would not be complete, as only frames $f_i \in \mathcal{F}^c_j$ could be used to infer the boats' position.
In the following, we present multiple solutions.

\subsubsection{Linear Interpolation}
The first, most simple solution is to linearly interpolate all boat positions $\mathbf{x}_{i, j}$ where $f_i \notin \mathcal{F}^c_j$. Interpolation for frame $f_i$ is based on two frames $f_l$ and $f_m$, $f_l, f_m \in \mathcal{F}^c_j$, where $l<i<m$, and $l$ and $m$ are chosen to be as close to $i$ as possible. In a more sophisticated approach, we aim to use optical flow to solve this problem.

\subsubsection{Optical Flow for Athlete Ordering}
For frame $f_i$, the number of detections $D$ in lane $j$ is not equal to $N$.
In this case, as it stands, we are not able to localize the boat tip.
However, let neighbouring frame $f_{i-1}\in F^c_j$ be a frame for which all detections are available and therefore the boat position is defined. 
From each athlete's image position, $\mathbf{a}'_{i-1,j, k}$ we estimate the optical flow vector $\mathbf{v}_{i-1 \rightarrow i, j, k}$ from frame $f_{i-1}$ to the following frame $f_i$ using Lucas-Kanade optical flow \cite{lucas_iterative_1981}.
Next, we match each position $\mathbf{a}'_{i-1,j, k} + \mathbf{v}_{i-1 \rightarrow i, j, k}$ to the nearest detection. This gives us $\mathcal{A}'_{i,j}$ and consequently $\mathcal{A}_{i,j}$. Note, that these are given their correct indices, which allows the corrective offsets to be applied, but some elements $\mathbf{a}_{i,j,k}$ may be empty due to missing detections.

\subsubsection{Optical Flow Interpolation}
Furthermore, we also implement interpolation based on the optical flow estimation.
For athlete positions in a frame $f_i \notin \mathcal{F}_c\left(j\right)$, if no detection is matched to the flow estimation, we simply use the estimation as the athlete's position, i.e. $\mathbf{a}'_{i,j, k} = \mathbf{a}'_{i-1,j, k} + \mathbf{v}_{i-1 \rightarrow i, j, k}$.
This ensures that all elements of $\mathcal{A}_{i,j}$ are not empty.
The described algorithm can be implemented to keep causality, only using flow information from a past to present frame, or breaking causality, using flow information in both time directions.

\subsection{Velocity Profile Estimation}

The boat position $\textbf{x}_j\left(t\right)$ is consequently filtered using an FIR filter, resulting in a robust distance-time-signal.
Velocity $\mathbf{v}_j(t)=\partial \mathbf{x}_j(t)/\partial t$ is estimated at every time step using second-order central difference approximation, and mapped to the distance signal for the velocity-distance profile.
Further, the signal is averaged in line with the GPS signal using a non-causal sliding window, centred around the discrete intervals of 12.5 meters.

\subsection{Estimation of Paddle Stroke Rate}

As previously mentioned, pacing strategies of canoe sprint races are not fully characterized by velocity profiles.
For a complete understanding of race strategy, stroke rate profiles must also be acquired.
To this end, we aim to reconstruct the paddle stroke rate for each boat using only the video footage.
We adopt a two stage approach. 
First, given the set $\mathcal{B}'_{i,j}$ of bounding boxes over the race distance for a lane $j$, we construct a one-dimensional motion signal $r_j\left(t\right)$ for each lane. To this end, we investigate two different methods which are described below. 
From this signal, we subtract the mean, pass through a Savitzky-Golay filter and extract a temporal stroke rate signal in the following manner. 

First, all local maxima are detected. Maxima within a specified tolerance (600ms for canoe, 300ms for kayak, based on known stroke frequency ranges for both disciplines) are merged to one maximum at the mean timestamp. This is repeated until no two peaks are within tolerance. Second, the time intervals between all detected peaks are calculated. Third, all interval lengths are averaged over $\pm2$ adjacent inter-peak-intervals. The interval periods are converted into stroke rates by dividing the interval length by the capture frame rate (fps) and converting to a frequency. The resulting values are mapped to a temporal signal, using the centre position between two subsequent local maxima as a timestamp. Finally, the stroke rate for any timestamp in between is linearly interpolated, which results in the temporal stroke rate signal $r_j(t)$. In order to reconstruct the stroke rate profile,  the temporal stroke rate signal is mapped to the race distance by synchronizing it with the boat's trajectory, $\mathbf{x}_j\left(t\right)$.

\begin{figure} 
    \centering
    \includegraphics[width=1.0\linewidth]{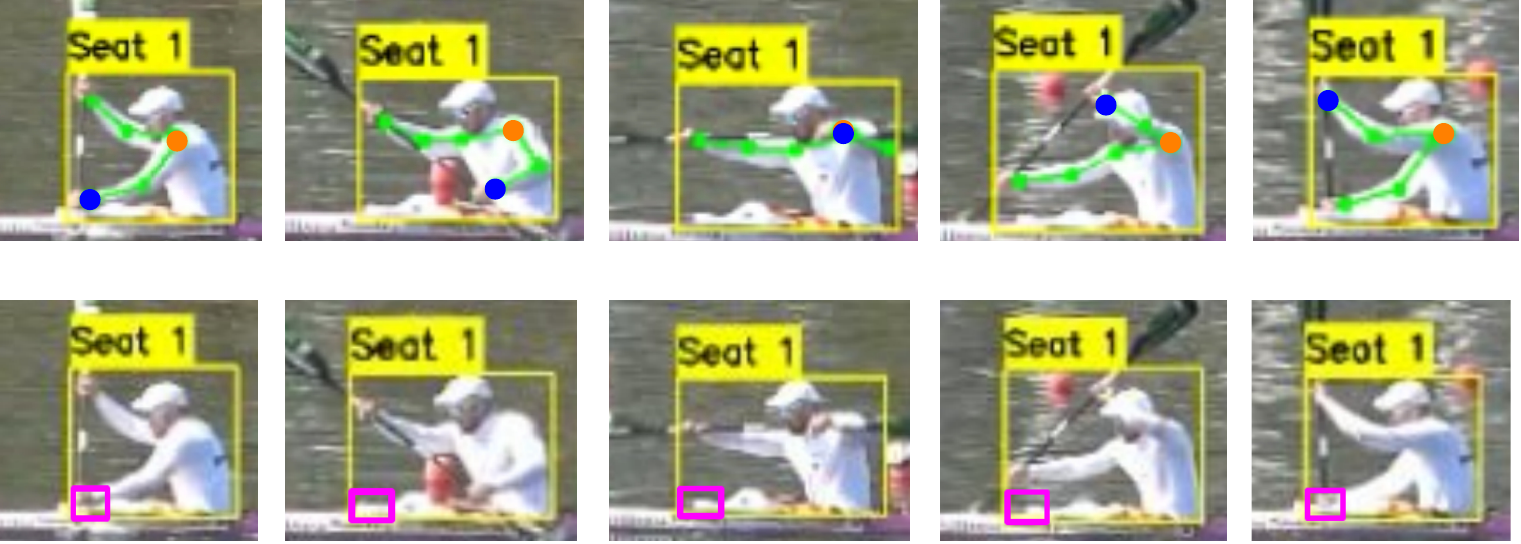}
    \caption{Visualization of stroke estimation methods across one paddle stroke.
    Top: A motion signal is built by calculating the Euclidean distance between the detected shoulder joint (orange) and wrist joint (blue) normalized by bounding box width in every frame.
    Bottom: The mean brightness of a patch 20 \% the width and height of the athlete bounding box (purple) in the left bottom corner is used as the motion signal.}
    \label{fig:stroke}
\end{figure}

As already mentioned, two methods are put forth to construct the motion signal $r_j\left(t\right)$. Both are visualized in Fig. \ref{fig:stroke}. First, a signal built using a ViTPose-H~\cite{xu_vitpose_2022} pose estimation model trained on the COCO dataset. 
Given the athlete bounding boxes, it returns predicted shoulder joints $\mathbf{s}'_{i,j}$ and wrist joints $\mathbf{w}'_{i,j}$ for each athlete. The component $r_{i,j}$ for the motion signal at frame $i$ is the Euclidean distance of the two joints, i.e. $r_{i,j} = ||\mathbf{s}'_{i,j} - \mathbf{w}'_{i,j}||$. Therefore, the motion signal $r_j(t)$ for an athlete on lane $j$ can be constructed by concatenating all components $r_{i,j}$.
Notably, we deploy the pretrained ViTPose model without domain specific fine-tuning or retraining. Our primary objective is to capture a motion signal and extract its fundamental frequency (the paddle stroke rate) over time, rather than to reconstruct the body poses in every frame. As occasional outliers and misdetections do not significantly impact the fundamental frequency, we do not perform frame-level validation of the joint predictions. In fact, end-to-end agreement of the predicted stroke rate and ground truth validates our method for signal extraction.

Second, we build a signal from the bounding box itself. The ROI described by the bounding box $\mathbf{b}'_{i,j}$ is greyscaled and blurred with an $11 \times 11$ kernel. Then, the component $r_{i,j}$ of the stroke rate signal is constructed using the mean brightness value in a small subregion in the lower left corner whose width and height are 20\% of the corresponding bounding box measures.
As before, the motion signal $r_j(t)$ for an athlete on lane $j$ can be constructed by concatenating all components $r_{i,j}$. Both methods of constructing the motion signal are evaluated in our Experiments, see Section \ref{sec:exp}.

For analysis of multi-athlete boat types, the motion signal is primarily derived from the front athlete, i.e. the athlete closest to the finish line.
In frames where this athlete was not detected, the nearest available detected athlete is chosen instead.

\subsection{Experiments}
\label{sec:exp}
We design a comprehensive suite of experiments designed to show our approaches capabilities. 
In the following, each experiment is outlined.

\subsubsection{U-Net Boat tip detection}

We evaluate the trained U-Net on the unseen test split of our boat tip dataset.
To assess performance, we determine mean and median pixel error and accuracy @ 1, 3, and 5 pixels respectively.

\subsubsection{Boat Measurement Validation}

To evaluate the consistency of our boat tip calibration, we collect measured offsets $\mathbf{o}_{j,k}$ from six K4 boats across 4 different nations, which are each present in two races of our dataset (both a heat and a semifinal or final). We compute the relative difference of measured offsets for each boat, i.e. $$\frac{1}{N}\sum_{k=1}^N \frac{||\mathbf{o}_{j,k}^{[1]} - \mathbf{o}_{j,k}^{[2]}||}{||\mathbf{o}_{j,k}^{[1]} + \mathbf{o}_{j,k}^{[2]}||/2}$$

The mean and median across all instances, is reported.

\subsubsection{Perspective Variance of Athlete Positioning Model}


As described earlier, we derive boat tip offsets for each athlete to be applied to each frame for accurate boat tip localization.
Through this we implicitly model the boat, more specifically the athlete spacings $$d_{n,m} = ||\mathbf{o}_{j,m} - \mathbf{o}_{j,n}||.$$
In this experiment, we aim to validate that this model holds over the full distance of the race under the perspective variance introduced by our panned and zoomed footage.
In each frame where all athletes are detected, we calculate athlete spacing based upon detected athlete positions 
$$\bar{d}_{n, m} = || \mathbf{a}_{i,j,m} - \mathbf{a}_{i,j,n}||.$$
Given these estimated athlete spacings, we determine the error to the offset based boat model (i.e., the athlete spacings as inferred in initial frames) for a particular boat using 
$$\frac{2!(N-2)!}{N!}\sum_{n=1}^N\sum_{m=n+1}^N \bar{d}_{n, m} - d_{n, m}.$$
We derive an error distribution over all multi-athlete data every 12.5m of the race distance and report the corresponding median and standard deviation for each distribution.

\subsubsection{Ablation of multi-athlete boat localization}

The introduction of team boats (K2/C2 and K4) necessitated a robust handling of boat localization, see Sec. \ref{sec:meth-multi}.

We ablate velocity profile accuracy using the described methods, i.e. linear interpolation, optical flow athlete ordering, and both causal and non-causal flow based interpolation. Only team boat types, i.e. C2, K2 and K4 are considered. We use $\textit{RMSE}$, $\textit{RRMSE}$ and $\textit{MAPE}$ to determine deviation of our estimations from the ground truth data on a per-sample basis.
Furthermore, we evaluate correlation using Spearman $\rho$.

\subsubsection{Evaluation of Velocity Profiles}

In this experiment, we compare all CV-based velocity profile results against the GPS ground truth velocity data.
For team boats, we use the best performing method for robust multi-athlete tracking.
The same metrics as in the previous experiment are used.


\subsubsection{Comparison of Stroke Rate Profile Estimation Methods}

Here, we evaluate the two methodologies for extracting stroke rate frequency from the video data against the gyroscopic ground truth data.
We again plot over distance travelled.
We quantify our results with $\textit{RMSE}$, $\textit{RRMSE}$ and $\textit{MAPE}$ at each distance increment, and  we show correlation of our results with ground truth using the Spearman $\rho$. Furthermore, we assess run-time of both methods, reported in seconds per video.

\subsubsection{Statistical Analysis of our Results}
\label{subsec:stats}

To further characterize our results, several complementary statistical measures and visual interpretation were used in order to assess our methods' performance. This applies to Sections \ref{subsec:comp_vel} and \ref{subsec:comp_stroke}.

In a first step, the distribution of the paired differences between both methods was examined to select appropriate analysis procedures. A Shapiro-Wilk test, alongside skewness and kurtosis assessment, indicates a non-gaussian distribution of our data. 
We therefore use non-parametric, distribution-free approaches. Results are reported as median [Q1, Q3].

We perform Bland-Altman analysis to assess overall agreement between our results and GPS/Gyro data. The Limits of agreement (LoA) were estimated empirically as the 2.5th and 97.5th percentiles of the differences, in accordance with the observed non-gaussian distribution. The corresponding 95\% confidence intervals were determined using bias-corrected, accelerated bootstrap resampling of the paired differences.

\subsubsection{Experimental Setup}

We implement our pipeline using Python 3.12, CUDA 13 and TensorFlow 2.12.
All experiments were run on a workstation with an Intel Core i7-8700k CPU and an NVIDIA RTX 4080 GPU.

\section{Results}

In this section, we present the results of the experiments outlined in the previous section.

\subsection{U-Net Boat tip detection}

First, we evaluate the trained U-Net on the unseen test split of the boat tip dataset.
The resulting performance metrics are presented in Table \ref{tab:boat-tip-accuracy}.
Median error between the predicted location and ground truth annotation was 1.72 pixels (mean of 2.97). Further evaluating accuracy shows 23.5 \% of samples are within 1 pixel of error, while 90.8 \% of all samples are within a 5 pixel tolerance.

\begin{table}[h]
\centering
\caption{Boat tip detection evaluation on test samples.}
\label{tab:boat-tip-accuracy}
\addtolength{\tabcolsep}{8pt}
\begin{tabular}{cc ccc}
\toprule
\multicolumn{2}{c}{\textbf{Pixel Error (px)}} & \multicolumn{3}{c}{\textbf{Accuracy (Acc@$k$) [\%]}} \\ 
\cmidrule(r){1-2} \cmidrule(l){3-5} 
\textbf{Mean} & \textbf{Median} & \textbf{1 px} & \textbf{3 px} & \textbf{5 px} \\ 
\midrule
2.97 & 1.72 & 23.5 & 81.6 & 90.8 \\ 
\bottomrule
\end{tabular}
\end{table}

\subsection{Boat Measurement Validation}

To further validate the calibration, we examine the offsets $\textbf{o}_1 .. \textbf{o}_{N}$ of 6 boats as measured across two distinct races.
We find a median relative difference of 2.65 \%, with a mean of 2.92 \%. This shows the independently calibrated boats are highly agreeable with one-another.

\subsection{Perspective Variance of Athlete Positioning Model}

\begin{figure}
    \centering
    \includegraphics[width=1.0\linewidth]{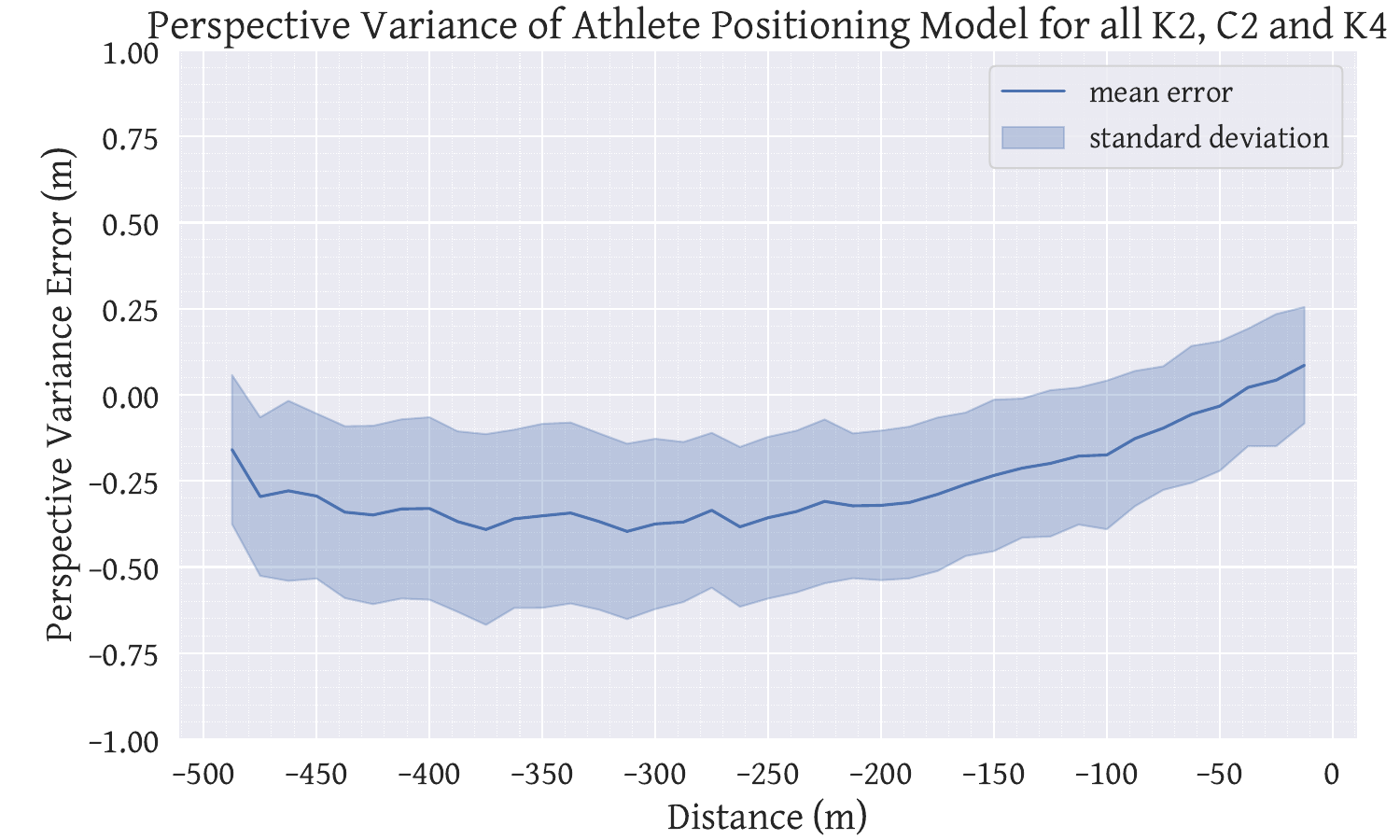}
    \caption{Perspective variance of the athlete positioning model relative to race distance. While a minor systematic negative shift in mean error occurs as the camera panning angle increases, the mean absolute error remains below 0.4 m. Standard deviation is consistent. The model maintains sufficient accuracy for boat tip estimation. Best viewed on screen.}
    \label{fig:e-over-distance}
\end{figure}

We show the results of the investigation of the error between modelled athlete spacings based on observations in initial frames (i.e., approximately perpendicular perspective) and estimated athlete spacings as observed over the course of the race. 
As can be seen in Fig. \ref{fig:e-over-distance}, the median error seems to undergo a small systematic shift towards negative values when the distance to the camera increases and therefore the panning angle of the camera changes accordingly. At around -400m, this shift vanishes slowly towards larger distances, while at the same time the standard deviation increases strongly. Importantly, the median absolute value of the measured error does not exceed 0.6 meters, which is an important indicator of how accurate the boat tip can be determined using the athlete offset model.

\subsection{Ablation of Tracking Method for Robust Multi-Athlete Boat Localization}

\begin{table}[htbp]
\caption{Robust multi-athlete tracking ablation. Details see text.}
\centering
\begin{minipage}{\textwidth}
\centering
\begin{tabular}{l|cccc}
\toprule
& \multicolumn{4}{c}{\textbf{Velocity Metrics} (Median [Q1 Q3])} \\
\cmidrule(lr){2-5}
\textbf{Method} & \textbf{RMSE} (m/s) & \textbf{RRMSE} & \textbf{MAPE} & \textbf{Spearman $\rho$} \\
\midrule
Linear Interpolation & 
  \makecell{0.444 \\{} [0.143 0.742]} & 
  \makecell{0.073 \\{} [0.029 0.150]} & 
  \makecell{0.053 \\{} [0.025 0.077]} & 
  \makecell{0.915 \\{} [0.882 0.944]} \\
\makecell[l]{Opt. Flow\\ Athlete Ordering} & 
  \makecell{0.433 \\{} [0.135 0.593]} & 
  \makecell{0.073 \\{} [0.026 0.107]} & 
  \makecell{0.053 \\{} [0.020 0.081]} & 
  \makecell{0.933 \\{} [0.816 0.945]} \\
\makecell[l]{Opt. Flow Interp.,\\ causal} & 
  \makecell{0.129 \\{} [0.063 0.170]} & 
  \makecell{0.022 \\{} [0.011 0.036]} & 
  \makecell{0.013 \\{} [0.007 0.022]} & 
  \makecell{0.968 \\{} [0.839 0.980]} \\
\makecell[l]{Opt. Flow Interp.,\\ non-causal} & 
  \makecell{\textbf{0.095} \\{} \textbf{[0.061 0.136]}} & 
  \makecell{\textbf{0.016} \\{} \textbf{[0.011 0.023]}} & 
  \makecell{\textbf{0.010} \\{} \textbf{[0.009 0.012]}} & 
  \makecell{\textbf{0.982} \\{} \textbf{[0.963 0.992]}} \\
\bottomrule
\end{tabular}
\label{tab:ma_abl}
\end{minipage}
\end{table}

Table \ref{tab:ma_abl} presents the results of the ablation study of tracking methods for multi-athlete boat localization.  It reveals a clear performance ranking across tracking methods. 
Replacing linear interpolation with optical-flow-based athlete ordering leads to negligible improvements if at all, increasing rank correlation from $\rho$ =  0.915 to 0.933.

A substantially larger gain is achieved by introducing optical-flow-based interpolation. 
The causal variant reduces \textit{RMSE} by nearly a factor of three compared to linear interpolation (0.129\,m/s vs.\ 0.433\,m/s) and improves correlation to $\rho = 0.968$.

The non-causal optical flow interpolation performs best overall, achieving the lowest \textit{RMSE} (0.095\,m/s) and \textit{MAPE} (0.010), as well as the highest correlation ($\rho = 0.982$). 

Overall, interpolation quality has a stronger impact than ordering alone, and leveraging non-causal temporal information yields the most accurate velocity estimates.

\subsection{Comparison of Velocity Profile with GPS}
\label{subsec:comp_vel}
\begin{table}[htbp]
\caption{Correlation and error measures for velocity profiles. Details see text.}
\centering
\begin{minipage}{\textwidth}
\centering
\begin{tabular}{l|cccc}
\toprule
& \multicolumn{4}{c}{\textbf{Velocity Metrics} (Median [Q1 Q3])} \\
\cmidrule(lr){2-5}
\textbf{Boat Type} & \textbf{RMSE} (m/s) & \textbf{RRMSE} & \textbf{MAPE} & \textbf{Spearman $\rho$} \\
\midrule
K1/C1 & 
  \makecell{0.079 \\{} [0.063 0.112]} & 
  \makecell{0.017 \\{} [0.014 0.023]} & 
  \makecell{0.011 \\{} [0.008 0.013]} & 
  \makecell{0.975 \\{} [0.960 0.985]} \\
K2/C2 & 
  \makecell{0.070 \\{} [0.048 0.155]} & 
  \makecell{0.015 \\{} [0.009 0.033]} & 
  \makecell{0.011 \\{} [0.006 0.017]} & 
  \makecell{0.965 \\{} [0.924 0.975]} \\
K4 & 
  \makecell{0.095 \\{} [0.061 0.136]} & 
  \makecell{0.016 \\{} [0.011 0.023]} & 
  \makecell{0.010 \\{} [0.009 0.012]} & 
  \makecell{0.982 \\{} [0.963 0.992]} \\ 
\midrule
All & 
  \makecell{0.078 \\{} [0.061 0.117]} & 
  \makecell{0.017 \\{} [0.013 0.024]} & 
  \makecell{0.011 \\{} [0.008 0.014]} & 
  \makecell{0.974 \\{} [0.949 0.984]} \\
\bottomrule
\end{tabular}
\label{tab:vel_rof}
\end{minipage}
\end{table}

The results of comparing GPS-derived velocity profiles with vision-based profiles as shown in Tab. \ref{tab:vel_rof} clearly indicate consistently high agreement across all boat classes. 
For K1/C1, our method achieves an \textit{RMSE}of 0.079\,m/s and a strong rank correlation ($\rho = 0.975$). 
Performance remains similarly robust for K2/C2 (RMSE: 0.070\,m/s, $\rho = 0.965$) and K4 (RMSE: 0.095\,m/s, $\rho = 0.982$).

While K2/C2 exhibits slightly higher variability, all boat types show low relative errors (\textit{MAPE} \~ 0.01) and strong monotonic agreement with GPS measurements.

Overall, the aggregated results across all classes (RMSE: 0.078\,m/s, \textit{RRMSE}: 0.017, \textit{MAPE}: 0.011,  $\rho = 0.974$) confirm that the proposed approach provides accurate and reliable velocity estimates independent of boat type.

\begin{figure}[htbp]
    \centering
    \subfigure[Bland-Altmann of velocity estimates.]{
        \includegraphics[width=0.47\textwidth]{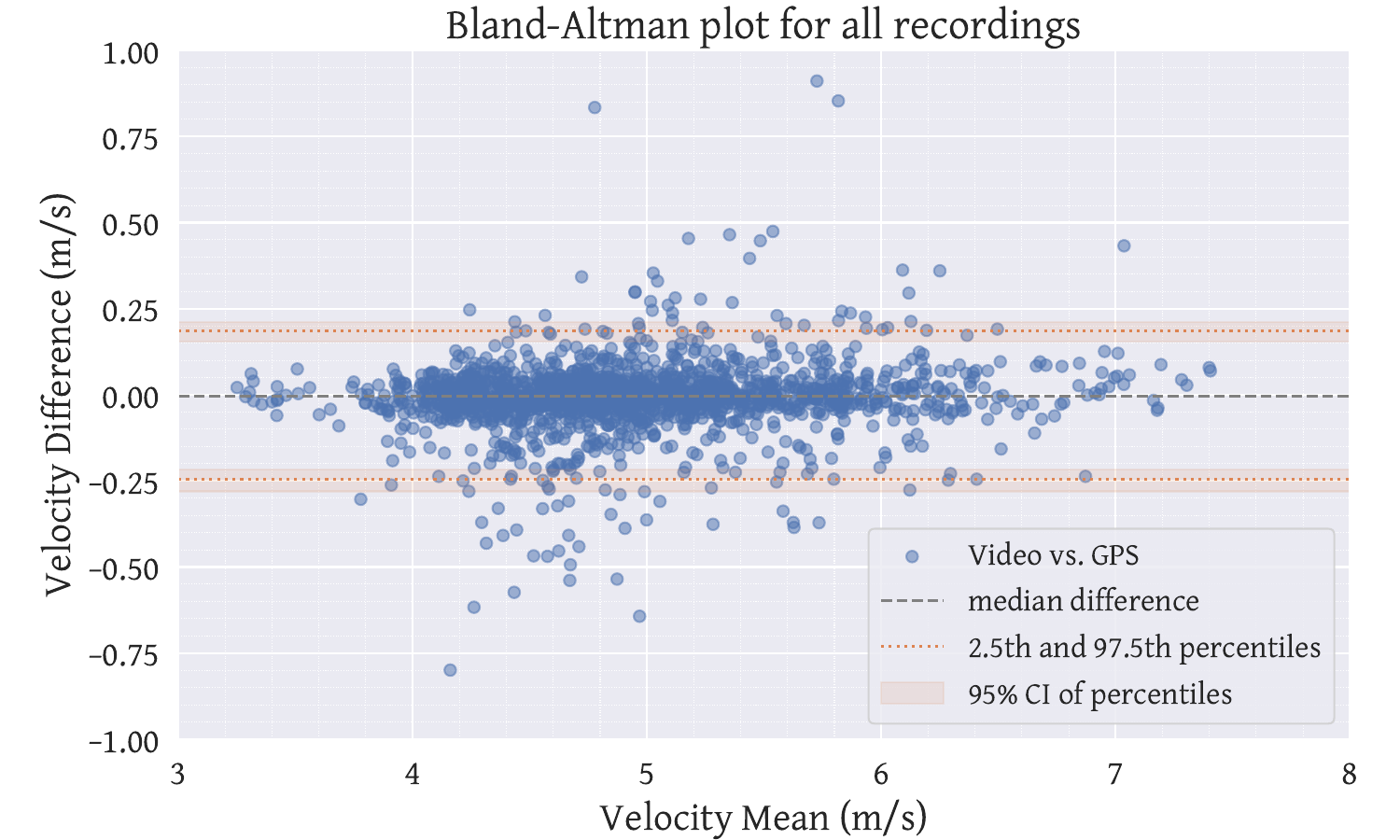}
        \label{fig:vel-bland}
    }
    \hfill
    \subfigure[Velocity Profile Error.]{
        \includegraphics[width=0.47\textwidth]{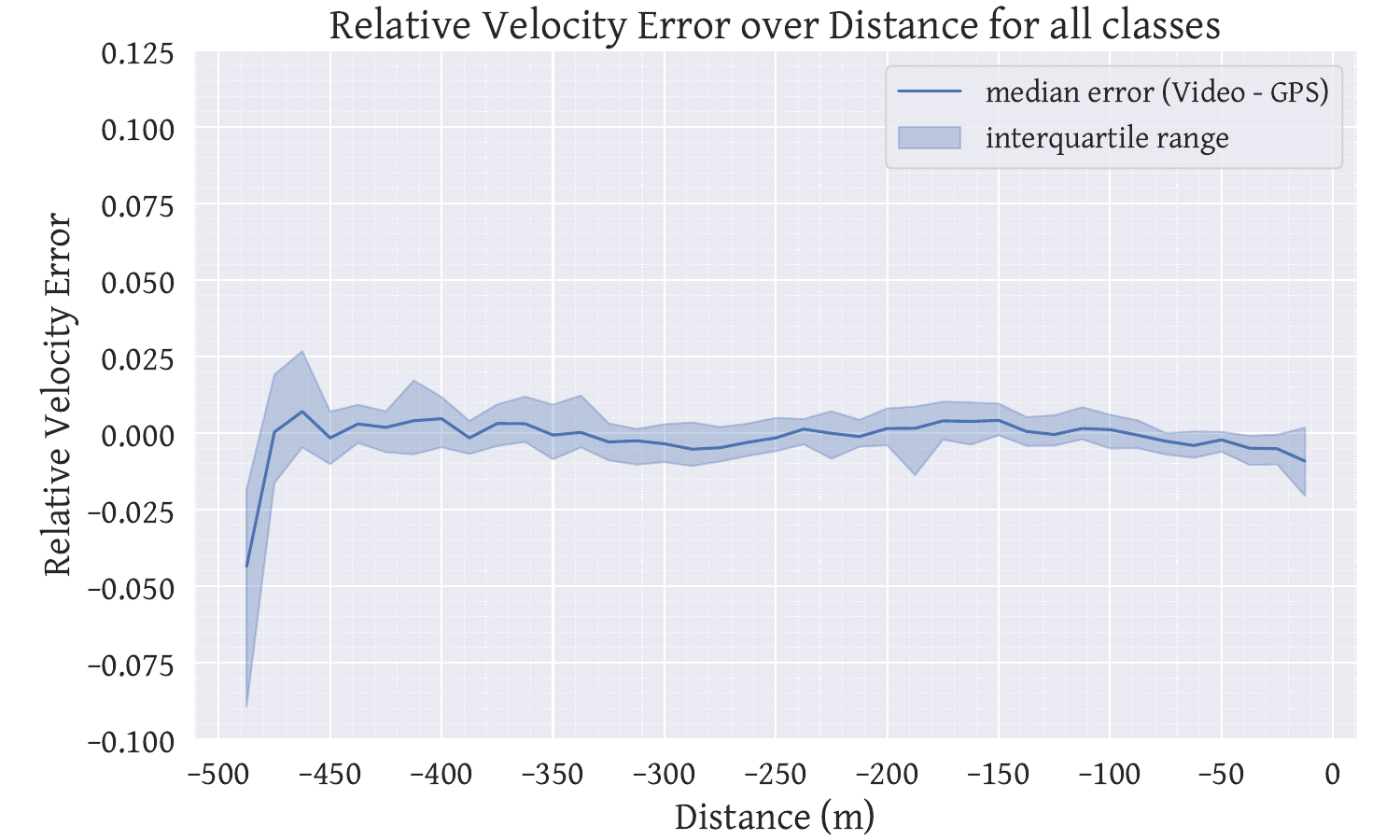}
        \label{fig:vel-dev}
    }
    \caption{Agreement analysis between the proposed velocity estimates and GPS ground truth. 
Left: Bland-Altman plot illustrating the median difference and limits of agreement (red, 2.5th and 97.5th percentiles), indicating minimal bias and only few outliers beyond the confidence bounds. Shaded area shows the confidence intervals of the limits of agreement, cmp. Sec. \ref{subsec:stats}.
Right: Relative velocity error over the full dataset, demonstrating low deviation from GPS measurements across the entire velocity range.}
\label{fig:res-vel}
\end{figure}

\begin{figure}
    \centering
    \subfigure{
        \includegraphics[width=0.47\textwidth]{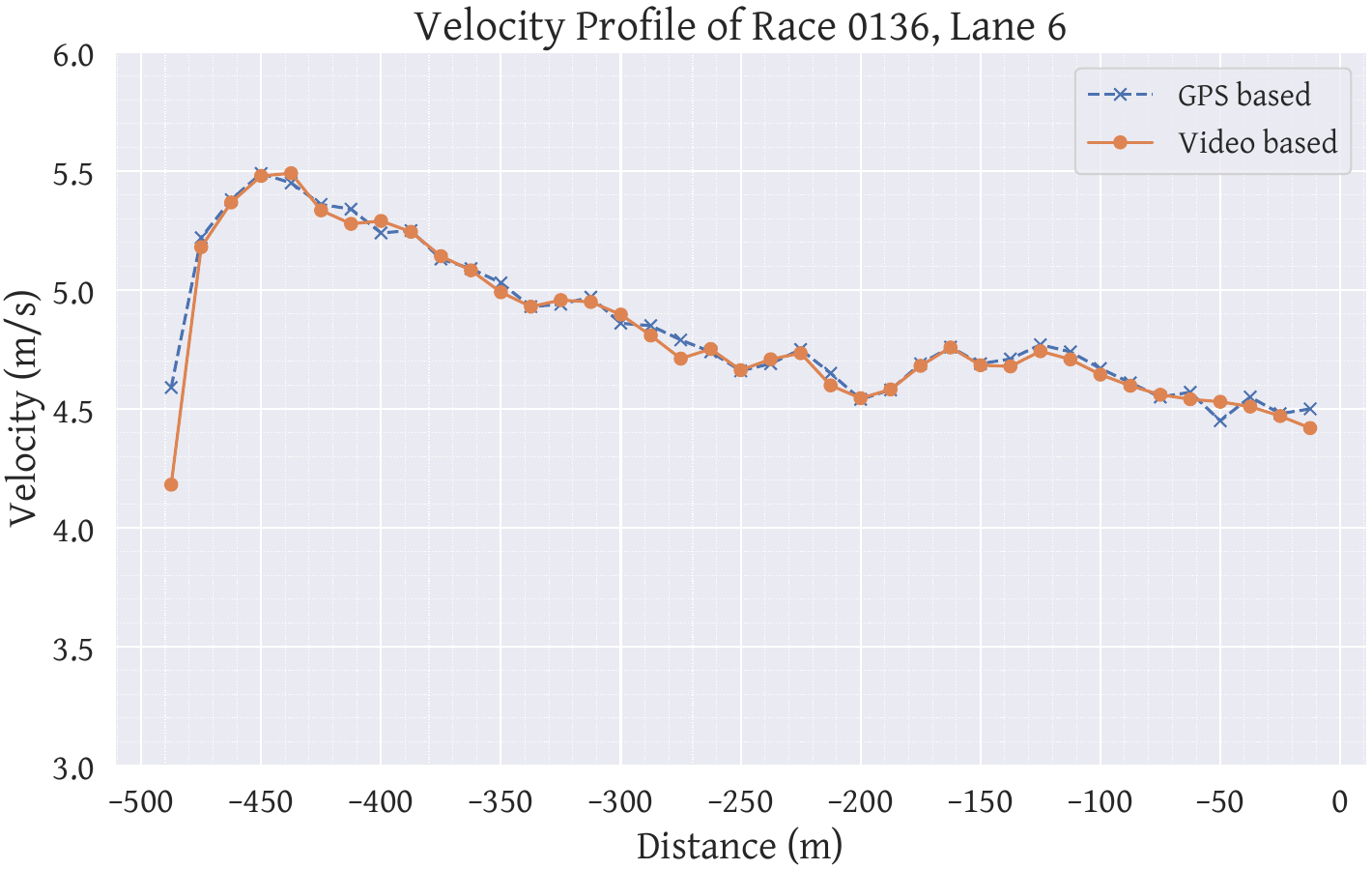}
        \label{fig:vel-prof1}
    }
    \hfill
    \subfigure{
        \includegraphics[width=0.47\textwidth]{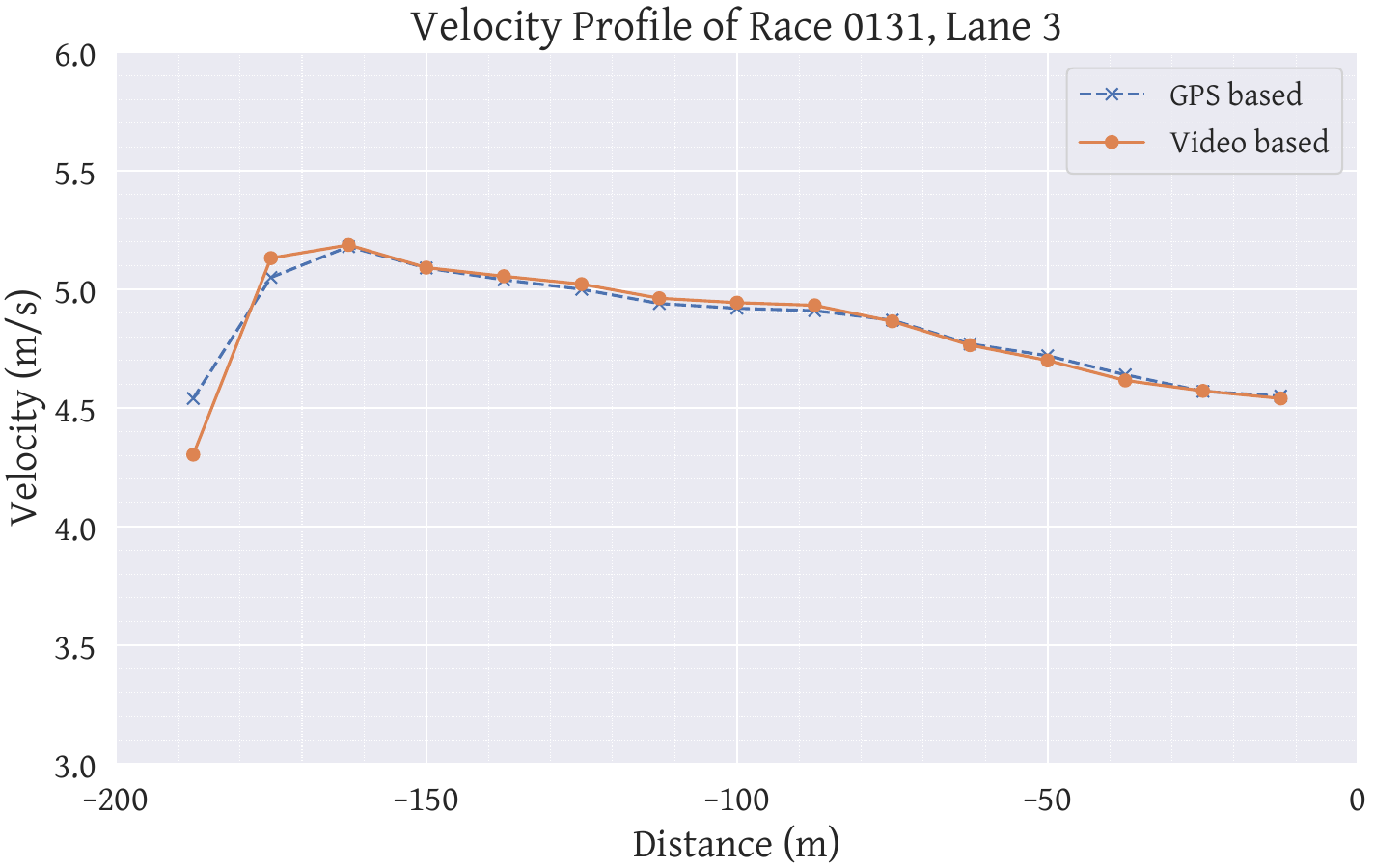}
        \label{fig:vel-prof2}
    }
    \caption{Results of GPS-based (blue) and video-based (orange) velocity profiles visualized over race distance. Left: 500m. Right: 200m. Both figures underpin the high agreement between video- and GPS-based measurements.}
    \label{fig:vel-profs}
\end{figure}

This is further supported by Fig. \ref{fig:res-vel}. Only a small minority of velocity estimates exceed the limits of agreement, and no systematic error can be observed at any velocity (Fig. \ref{fig:vel-bland}). Over all data, the velocity profile shows little relative deviation from ground truth GPS data, see Fig. \ref{fig:vel-dev}.
In Figure \ref{fig:vel-profs} we show two example velocity profiles.

\subsection{Comparison of Stroke Rate Methods}
\label{subsec:comp_stroke}

\begin{table}[h]
\centering
\caption{Performance metrics and computational efficiency for stroke rate estimation.}
\label{tab:stroke-metrics}
\begin{tabular}{@{}lcccc@{}}
\toprule
 & \multicolumn{4}{c}{\textbf{Stroke Rate} (Median [Q1 Q3])}\\ 
\cmidrule(lr){2-5}
\textbf{Method} & \textbf{RMSE} (bpm) & \textbf{RRMSE} & \textbf{MAPE} & \textbf{Spearman $\rho$}\\ \midrule
\textbf{ViTPose} & \makecell{\textbf{1.753} \\{} \textbf{[0.868 2.921]}} & \makecell{\textbf{0.014} \\{} \textbf{[0.009 0.022]}} & \makecell{\textbf{0.009} \\{} \textbf{[0.006 0.013]}} & \makecell{\textbf{0.975} \\{} \textbf{[0.921 0.990]}}\\
\textbf{BBox}    & \makecell{6.467 \\{} [4.399 10.031]} & \makecell{0.070 \\{} [0.039 0.088]} & \makecell{0.039 \\{} [0.019 0.053]} & \makecell{0.807 \\{} [0.590 0.914]}\\ \bottomrule
\end{tabular}
\end{table}

\begin{figure}[htbp]
    \centering
    \subfigure[Bland-Altmann of stroke rate estimates.]{
        \includegraphics[width=0.47\textwidth]{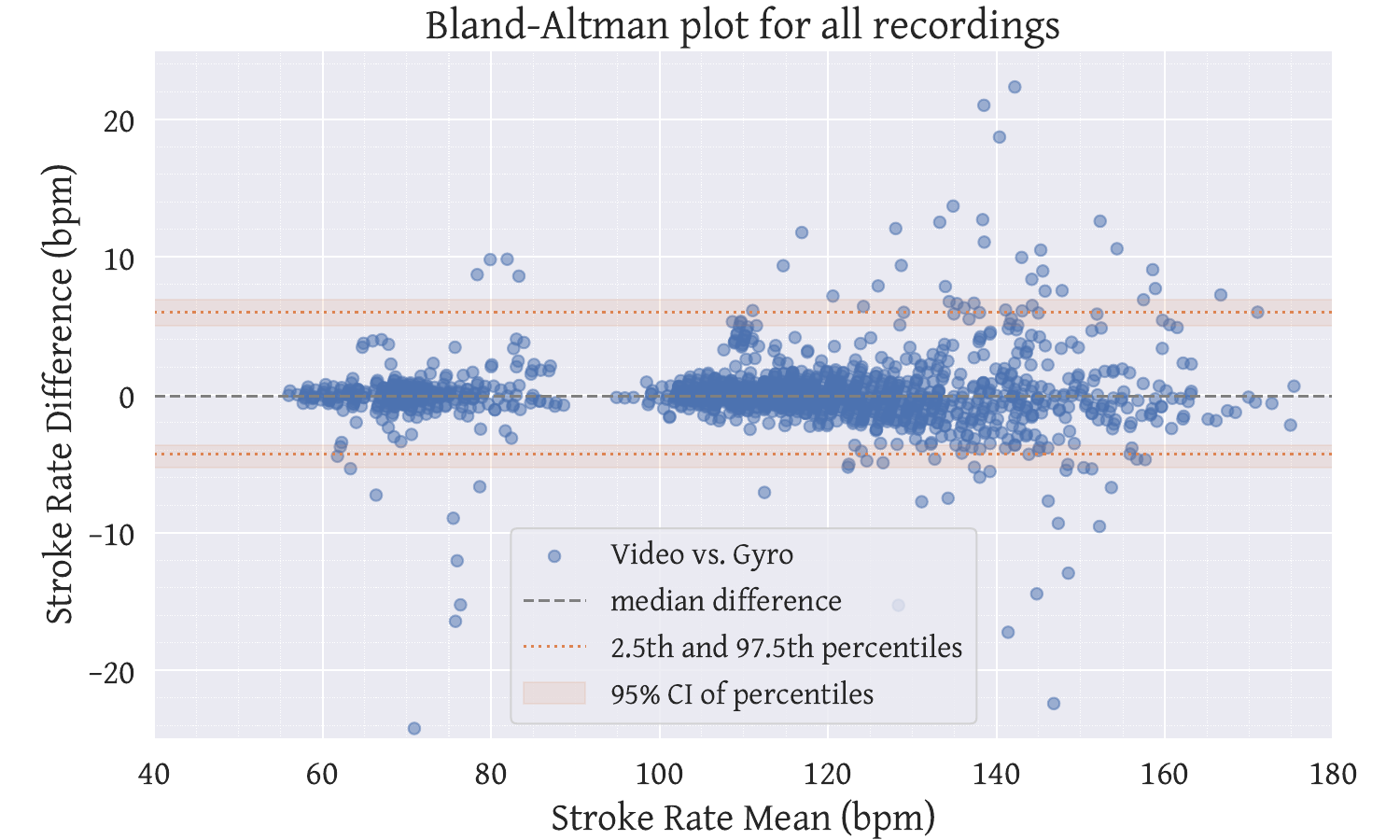}
        \label{fig:stroke-bland}
    }
    \hfill
    \subfigure[Stroke Rate Profile Error.]{
        \includegraphics[width=0.47\textwidth]{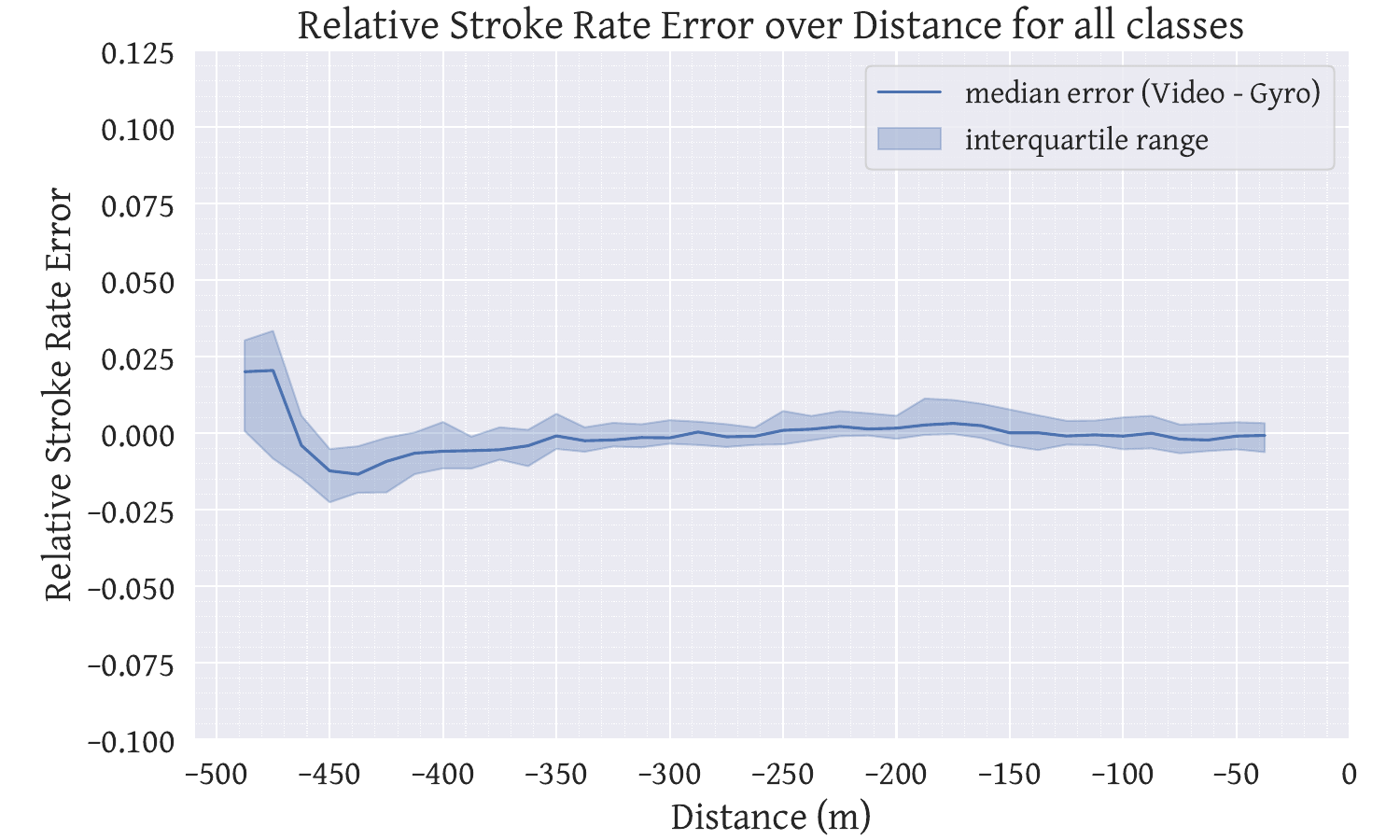}
        \label{fig:stroke-dev}
    }
    \caption{Agreement analysis between ViTPose stroke rate estimates and Gyrometric ground truth. 
Left: Bland-Altman plot illustrating the median difference and limits of agreement (2.5th and 97.5th percentiles), indicating minimal bias and only few outliers beyond the confidence bounds. Shaded area shows the confidence intervals of the limits of agreement, cmp. Sec. \ref{subsec:stats}. The two distinct clusters are caused by the substantially different stroke rates of canoe and kayak athletes.
Right: Relative stroke rate error over the full dataset, demonstrating low deviation from Gyrometric measurements across the entire stroke rat range. The error range slightly varies over distance which can be attributed to occlusions that consistently occurred (cmp. Fig.~\ref{fig:okkls}) in the stroke rate validation dataset.}
\label{fig:res-stroke1}
\end{figure}

\begin{figure}[htbp]
    \centering
    \subfigure[Occlusion at approximately 120m race distance.]{
        \includegraphics[width=0.47\textwidth]{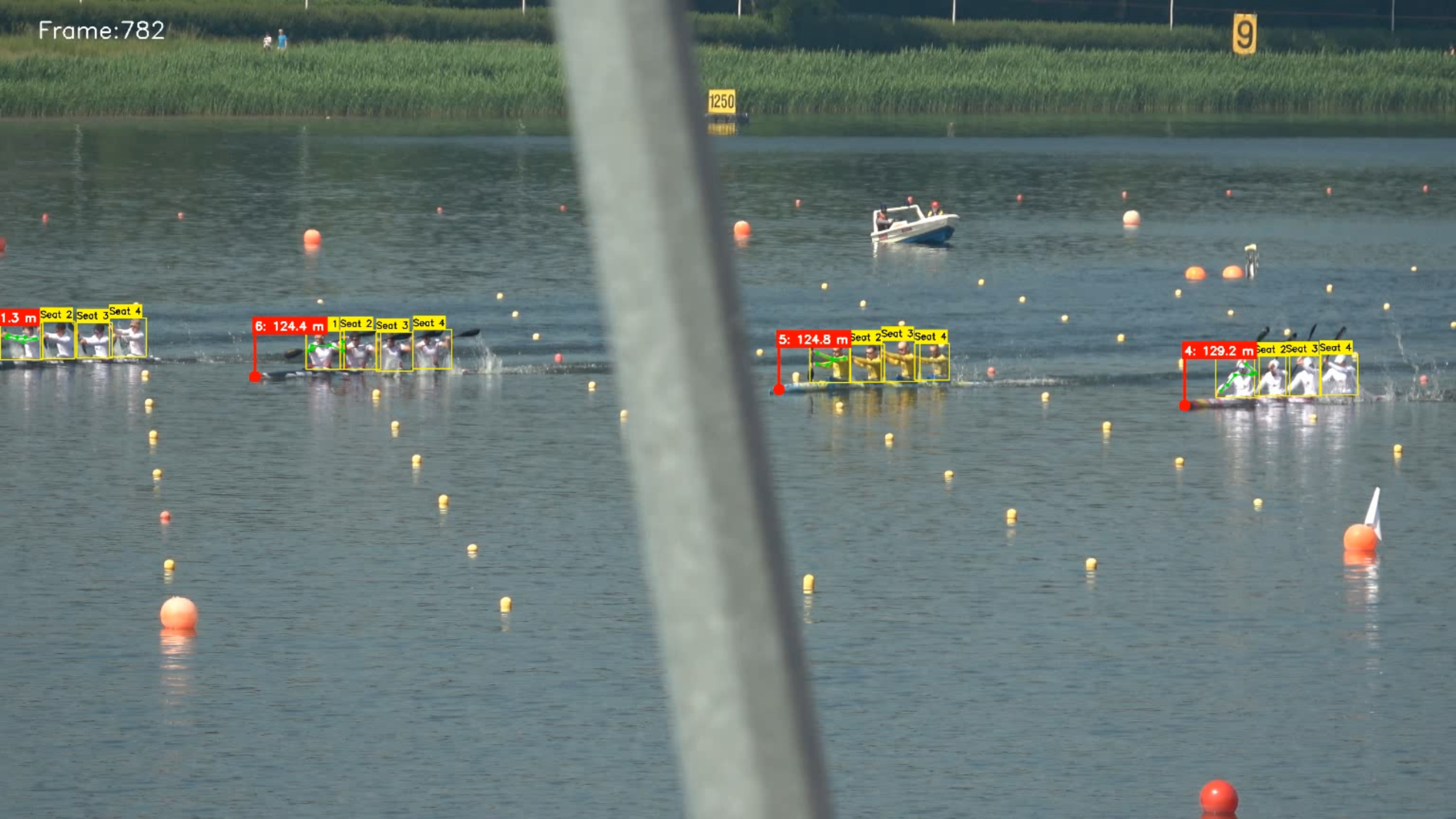}
        \label{fig:okkl120}
    }
    \hfill
    \subfigure[Occlusion at approximately 320m race distance.]{
        \includegraphics[width=0.47\textwidth]{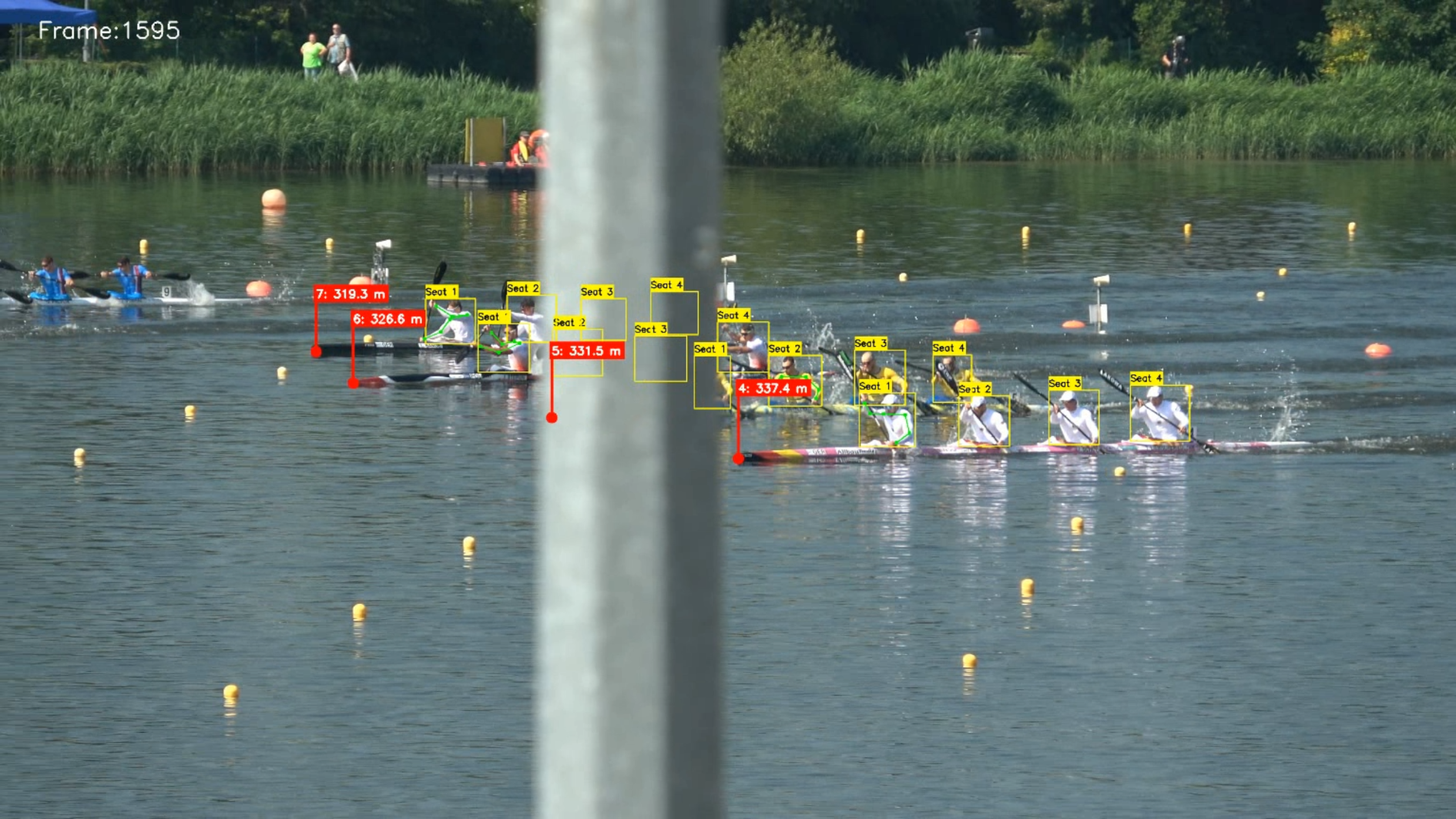}
        \label{fig:okkl320}
    }
    \caption{Stroke rate analysis was conducted over all boat classes, but data for multi-athlete boats consistently suffered from occlusions due to several poles that become visible while moving the camera to follow the boats. The occlusion occurred at around 120m (left) and 320m (right) race distance. These correspond to the higher standard deviations observed in our predictions at -380m and -180m, see Fig.~\ref{fig:res-stroke1}.}
\label{fig:okkls}
\end{figure}

We evaluate two distinct approaches for modelling a motion signal $r(t)$ for paddle stroke estimation. The results for both approaches are summarized in Tab. \ref{tab:stroke-metrics}. 
The ViTPose-based approach clearly outperforms the method relying solely on bounding box data as the motion signal. We show Bland-Altman analysis and relative deviation over distance of this method in Fig. \ref{fig:res-stroke1}. Note that the distribution of the relative deviation across distance is subject to some variations which arise due to systematic occlusions in the video data as shown in Fig. \ref{fig:okkls}.

Using ViTPose, we obtain a sample-to-sample $\textit{RMSE}$ of 1.753 m/s and a $\textit{MAPE}$ of 0.009. In contrast, the bounding box signal yields an $\textit{RMSE}$ of 6.467 m/s and a $\textit{MAPE}$ of 0.039. For both metrics, the error of the ViTPose-based method is less than one third of that observed with the bounding box approach.

\begin{figure}[htbp]
    \centering
    \subfigure[]{
        \includegraphics[width=0.47\textwidth]{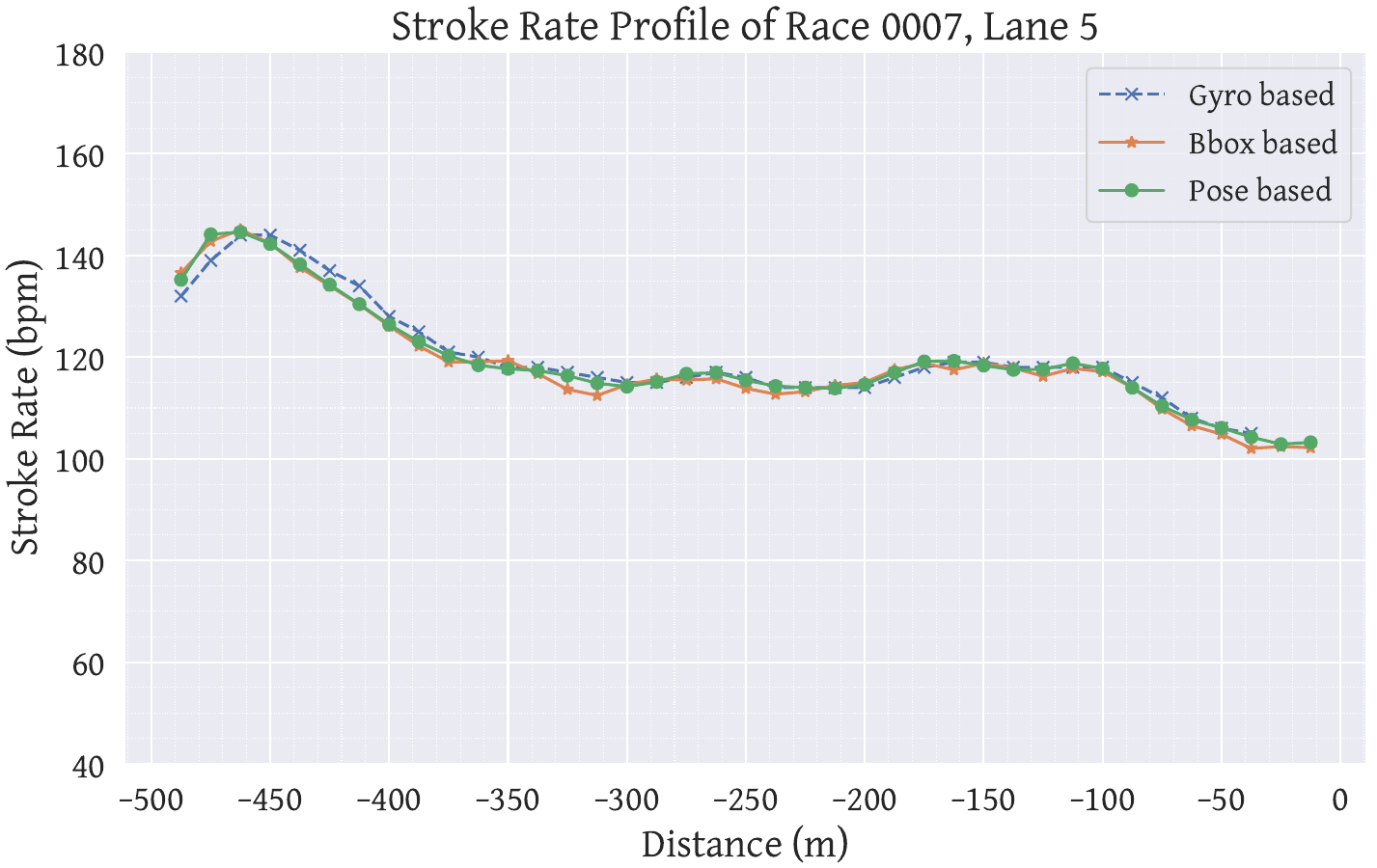}
        \label{fig:good_bbox_stroke}
    }
    \hfill
    \subfigure[]{
        \includegraphics[width=0.47\textwidth]{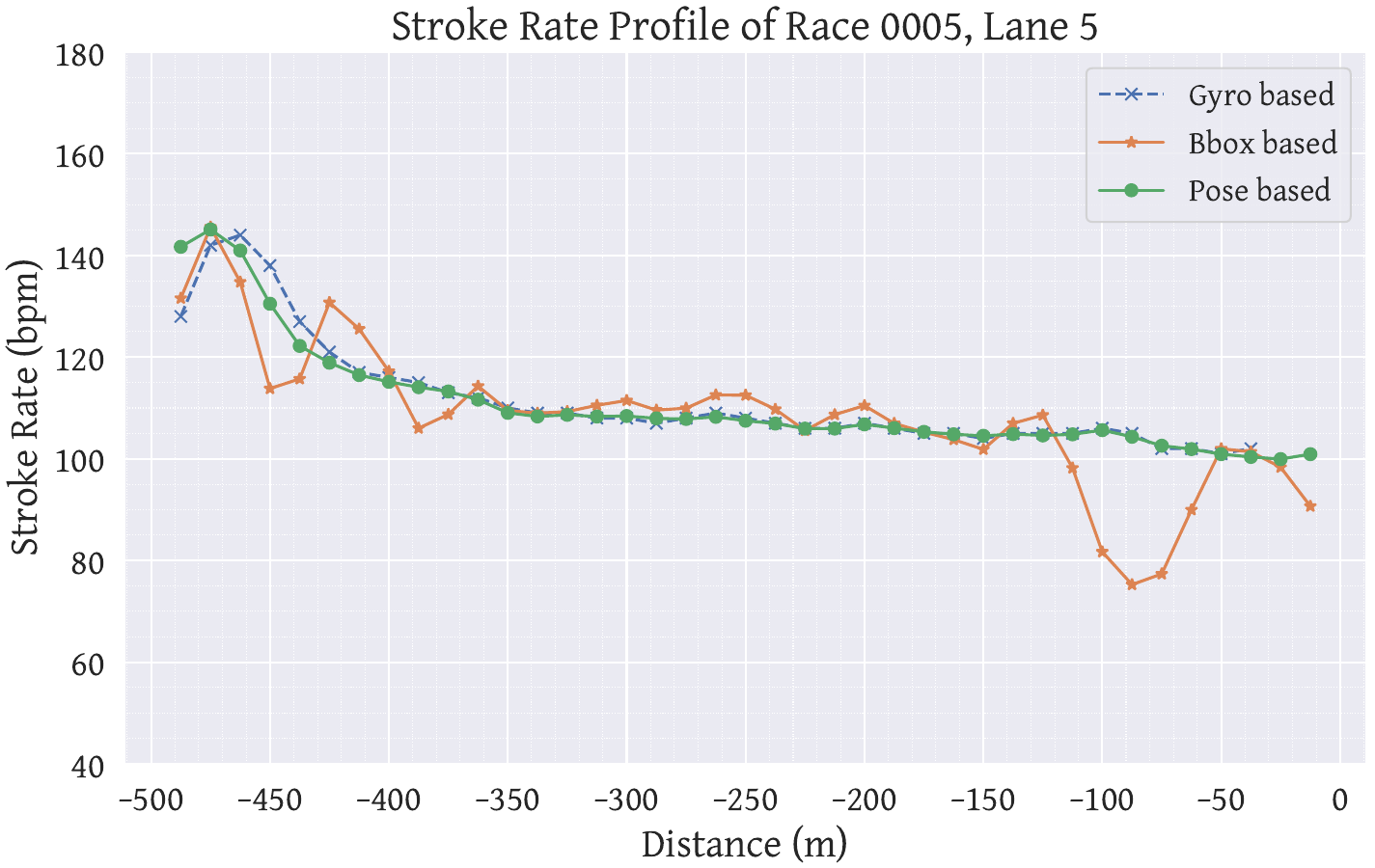}
        \label{fig:bad_bbox_stroke}
    }
    
    \caption{Results of both Pose-based (green) and Bounding Box based (orange) stroke rate estimation, visualized for two races. The left race shows both methods performing well, as they are both able to capture the form of the profile. On the right, both methods show weaknesses and do not correlate exactly with ground truth data. However, the bounding box method delivers much worse results.}
    \label{fig:res-stroke}
\end{figure}

Moreover, the ViTPose-derived profile exhibits substantially stronger correlation with the reference signal compared to the bounding box-based profile (Spearman $\rho$ of 0.975 vs.\ 0.807). A qualitative comparison for two races is shown in Fig.~\ref{fig:res-stroke}.

In terms of computational cost, reconstructing the stroke rate for all lanes in a video requires on average 1179 seconds using ViTPose, whereas the bounding box-based method requires only 411 seconds. For context, homography estimation across all frames takes 2243 seconds on average, while velocity profile estimation requires 537 seconds of computation time.

\section{Discussion}

\begin{figure}[t!]
    \centering
    \includegraphics[width=1.0\linewidth]{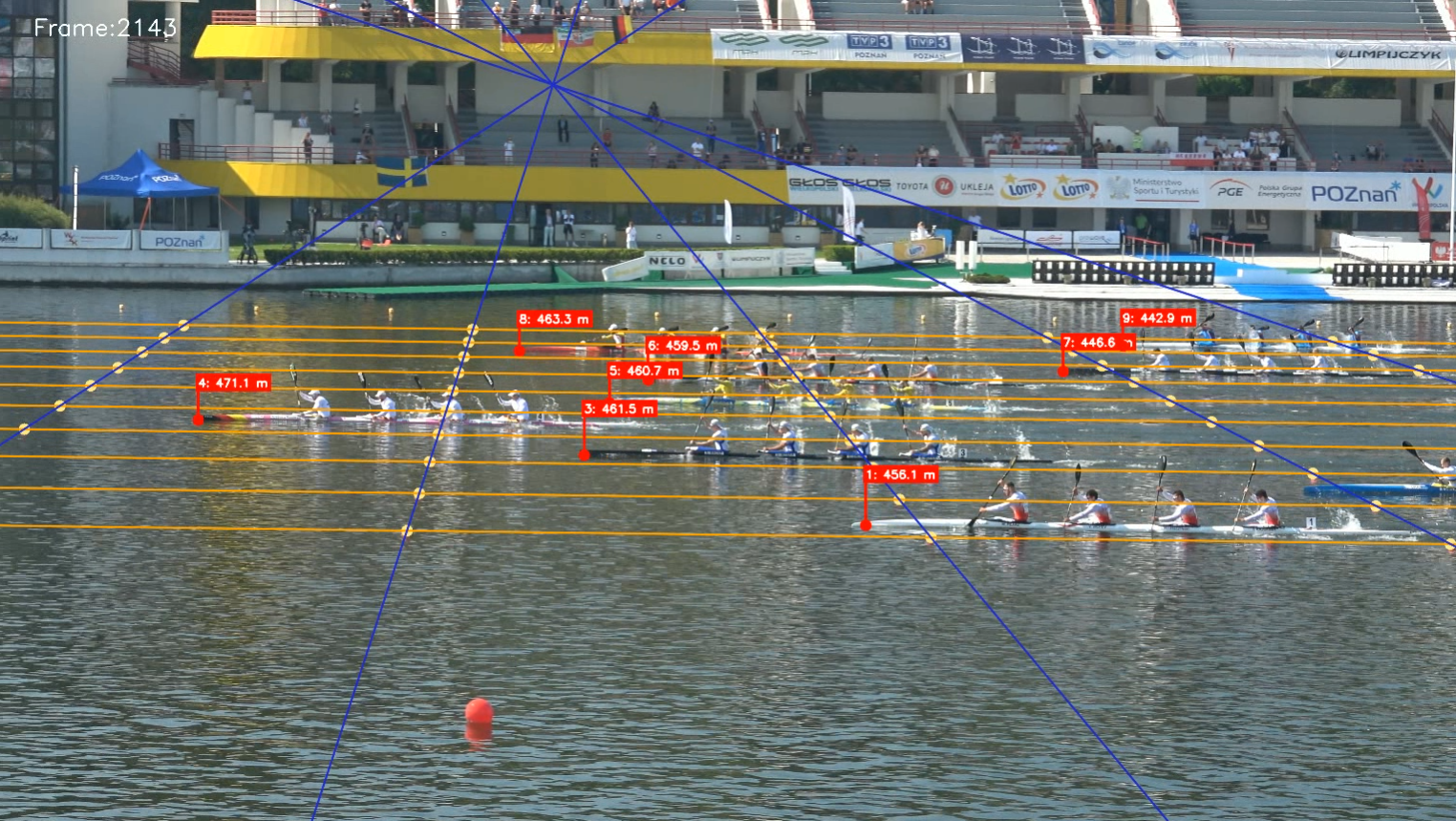}
    \caption{Demonstration of the application of the methodological extensions to analyse team boats as presented in this paper to a K4 canoe race over 500m distance. See. Fig.~\ref{fig:K1_01} for annotations. Best viewed on screen.}
    \label{fig:K4_01}
\end{figure}

In this work, we substantially extend prior work on automatic analysis of canoe sprint races \cite{matthes_reconstructing_2025} towards a unified, video-only system capable of handling multi-athlete boat classes and estimating both velocity and stroke rate profiles.
The methodology was revised to generalize from single-athlete boats to C2, K2 and K4, while maintaining performance for C1 and K1.
Central to the extension from single- to multi-athlete boats is a learned boat tip calibration using a U-Net model, enabling individual offset estimation per athlete instead of relying on a static empirical offset, and the propagation of the athlete order which is essential to selecting the correct offset for each athlete in a boat. Recovering the order of athletes in each frame finally allows the recovery of the boat location in each frame of a video (cmp. Fig.~\ref{fig:K4_01}).

The U-Net achieves strong localization performance on the held-out test split (Acc@5 px of 90.8\%), indicating that in the vast majority of frames the boat tip is detected within a 5-pixel tolerance.
In addition, comparing calibrated offsets of identical boats across independent race recordings reveals only very small deviations (2.7\%), further supporting the stability and reproducibility of the learned calibration.
Together, these findings suggest that the automated boat tip estimation is both accurate and consistent across different videos and recording conditions.

When analysing the modelled offset error over the course of a race, a slight negative bias emerges.
It can be interpreted as a small systematic localization error that is introduced when estimating the boat tip based on the athletes positions and using the calibrated offsets.
However, this bias evolves only gradually over distance and does not exhibit abrupt fluctuations.
The resulting position signal (i.e., the boat position over time) is used (1) to derive a velocity signal and (2) to map estimated velocity and stroke rate at each time step to a position on the track, i.e. to the race distance.
This leads to the question of how this bias might impact velocity estimation and position mapping.

Since velocity estimation depends on temporal differences rather than absolute position alone, such a slowly varying bias does not meaningfully distort the reconstructed velocity profiles.
This is corroborated by the strong agreement between estimated and ground truth velocities, indicating that the residual model error remains practically negligible for tactical analysis.

While a negative bias exists at large distances, it is a function of perspective. Because the perspective remains nearly constant between two distant points (e.g., 312.5m and 300.0m), the bias is locally stationary. Consequently, the mapping function remains internally consistent, which is validated by the high correlation ($\rho \approx 0.96+$) between our estimated profiles and ground truth.

Another key contribution is the use of optical flow for multi-athlete tracking when detections fail.
Using optical flow to successfully order detections gives modest gains over the baseline, while interpolation via optical flow for missing detections improves estimated velocities significantly.
Non-causal use of flow shows a measurable performance increase, which needs to be kept in mind for real-time applications in further studies.

Estimated velocity profiles exhibit low relative errors ($\textit{MAPE} \approx 0.01$) and high correlations with ground truth ($\rho \approx 0.96+$) across all boat classes.
No significant performance shift between boat types was observed, indicating that the geometric modelling and calibration strategy generalize well across boat types. Although the boat class C4 is not contained in our dataset and therefore not part of the presented analysis, there is reason to believe that our methodology generalizes for this class as well.
Larger differences only appeared during the first 25–50 metres of races. This discrepancy likely stems from GPS inaccuracies during high-acceleration phases, a phenomenon documented in previous validation studies~\cite{fernandes_validation_2024}. While GPS sensors often suffer from positional lag and sampling jitter during the explosive start of a race, our video-only system maintains high temporal resolution and visual consistency. Given that our detections remain stable and the resulting trajectory aligns with the physical constraints of the boat's motion, we believe these differences are indicative of GPS measurement error rather than a failure in our vision-based tracking
The accuracy achieved lies within practically relevant margins for tactical race analysis.

For stroke rate estimation, we compared a ViTPose-based pose estimation approach with a method relying solely on bounding box signals.
The ViTPose-based method achieves an error of approximately 2 bpm and a correlation of $\rho = 0.975$ to gyroscopic ground truth, clearly outperforming the bounding box approach ($\textit{RMSE} = 6.467$, $\rho = 0.807$).
However, ViTPose increases computational cost for stroke estimation by 187\%.
When considering the complete processing pipeline, this corresponds to an overall increase of 24\%, which appears justified given the substantial accuracy gain.
Nevertheless, the bounding box method remains attractive for low-resource or near-real-time settings.
Its main limitation lies in its sensitivity to occlusions and the relatively weak motion signal extracted from a single region. 

Several limitations remain.
Homography initialization still requires manual selection and identification of four non-coplanar buoys. However, automating this task should be a straightforward addition to our method.
Frame-wise homography estimation ignores the inherent temporal smoothness condition of a video-sequence, which could lead to faulty transformations to real-world coordinates. Such cases do not lead to significant errors, as our measurements are ultimately collected over many frames.
Optical flow may fail under severe occlusions, and pose estimation can degrade under motion blur. In practice, these should occur in only insignificant timeframes, and should therefore be easily compensated by interpolation.
Our method is based on a strong assumption of the buoy-based scene geometry. Imperfect buoy placements or minor buoy movement due to water motion might introduce inaccuracies into homography estimates. These effects however are marginalized both spatially when estimating a homography from a larger set of buoys and temporally as the estimated positions are smoothed over frames.
The boat positions measurement is offset-based using the detected athlete positions. This implies a reliable estimate over the course of the race. Our method is prone to systematic shift of this offset, which occurs due to the change of perspective during the race. At the start, the perspective corresponds to a rather frontal viewpoint, moving toward a sagittal perspective during the race. Given some subsets of frames, however, the perspective is approximately constant. Hence, this perspective change might introduce a very low frequent component in the velocity signal. Our results emphasize that this does not affect the accuracy of our method.
However, we strongly believe our results emphasize the generalizability of our approach. Although minor inaccuracies can occur, the strong end-to-end agreement suggests these are not significant.

Future work should address temporal smoothing or joint optimization of homographies to reduce frame-to-frame instability.
For stroke rate estimation, the bounding box signal could be further refined, for instance by modelling differential motion between multiple regions or applying dimensionality reduction techniques such as PCA to extract stronger periodic components.
Extensions towards para-disciplines and transfer to rowing appear feasible given the geometric generality of the approach.
Finally, enabling fully causal, real-time estimation would require a fundamentally revised homography estimation strategy and stricter constraints on tracking and calibration, but represents a promising direction for applied competition analysis.

\bibliographystyle{ieeetr}
\bibliography{references}

\renewcommand{\indexname}{Author Index}
\printautindex

\end{document}